\documentclass[10pt,twocolumn,letterpaper]{article}

\usepackage{cvpr}              
\usepackage{caption}

\usepackage{xcolor}
\usepackage{mathtools}
\usepackage{multirow}
\usepackage{graphicx}
\usepackage{subcaption}   
\usepackage[table,xcdraw]{xcolor}
\usepackage{svg}
\usepackage{wrapfig}

\definecolor{darkgreen}{RGB}{0,100,0}
\newcommand{\appref}[1]{Supp.~\ref{#1}}

\usepackage{pifont}
\definecolor{DarkGreen}{RGB}{17,193,56}
\newcommand{\cmark}{\textcolor{DarkGreen}{\ding{51}}}%
\newcommand{\xmark}{\textcolor{red}{\ding{55}}}%

\definecolor{cvprblue}{rgb}{0.21,0.49,0.74}
\definecolor{cvprpink}{RGB}{230,0,115}  
\usepackage[pagebackref,breaklinks,colorlinks,
            linkcolor=cvprblue,
            citecolor=cvprblue,
            urlcolor=cvprpink]{hyperref}

\def\paperID{} 
\def\confName{CVPR}
\def\confYear{2026}

\title{MoD-DPO: Towards Mitigating Cross-modal Hallucinations in Omni LLMs using Modality Decoupled Preference Optimization}

\author{Ashutosh Chaubey, Jiacheng Pang, Mohammad Soleymani\\
University of Southern California\\
{\tt\small \{achaubey, pangj\}@usc.edu, soleymani@ict.usc.edu}
}

\begin{document}

\twocolumn[{
\maketitle
\vspace{-3em}
\begin{center}
{\large
\href{https://mod-dpo.github.io}{Project Page: mod-dpo.github.io}
}
\end{center}
\vspace{-1em}
}]

\begin{abstract}
Omni-modal large language models (omni LLMs) have recently achieved strong performance across audiovisual understanding tasks, yet they remain highly susceptible to cross-modal hallucinations arising from spurious correlations and dominant language priors. In this work, we propose Modality-Decoupled Direct Preference Optimization (MoD-DPO), a simple and effective framework for improving modality grounding in omni LLMs. MoD-DPO introduces modality-aware regularization terms that explicitly enforce invariance to corruptions in irrelevant modalities and sensitivity to perturbations in relevant modalities, thereby reducing unintended cross-modal interactions. To further mitigate over-reliance on textual priors, we incorporate a language-prior debiasing penalty that discourages hallucination-prone text-only responses. Extensive experiments across multiple audiovisual hallucination benchmarks demonstrate that MoD-DPO consistently improves perception accuracy and hallucination resistance, outperforming previous preference optimization baselines under similar training budgets. Our findings underscore the importance of modality-faithful alignment and demonstrate a scalable path toward more reliable and resilient multimodal foundation models.
\end{abstract}    
\section{Introduction}
\label{sec:intro}

Recent progress in multimodal (or \emph{``omni"}) large language models has rapidly expanded language-only reasoning into rich perceptual settings that span images, audio, and video \cite{Qwen2.5-VL,xu2025qwen25omnitechnicalreport,omnivinci2025,ghosh2025audioflamingo3_af3,feng2025videor1}. Systems such as OmniVinci \cite{omnivinci2025} demonstrate that carefully designed alignment modules, temporal representations, and curated omni-modal dialogue data can unlock strong cross-modal understanding at competitive efficiency. At the same time, foundation series like Qwen2.5 Omni \citep{xu2025qwen25omnitechnicalreport} have shown that scale, post-training via DPO \cite{rafailov2023direct_dpo}/GRPO \cite{shao2024deepseekmath_grpo}, and tool-augmented vocabularies translate into broadly useful capabilities and robust instruction following across modalities. Beyond single-turn perception, emerging omni LLMs aim for unified, interactive audiovisual-language reasoning, pointing toward agents that both \emph{see} and \emph{listen} before they \emph{think and respond} \citep{tong2025interactiveomni}.

\begin{figure}
    \centering
    \includegraphics[trim={0 2mm 0 0},clip,width=\linewidth]{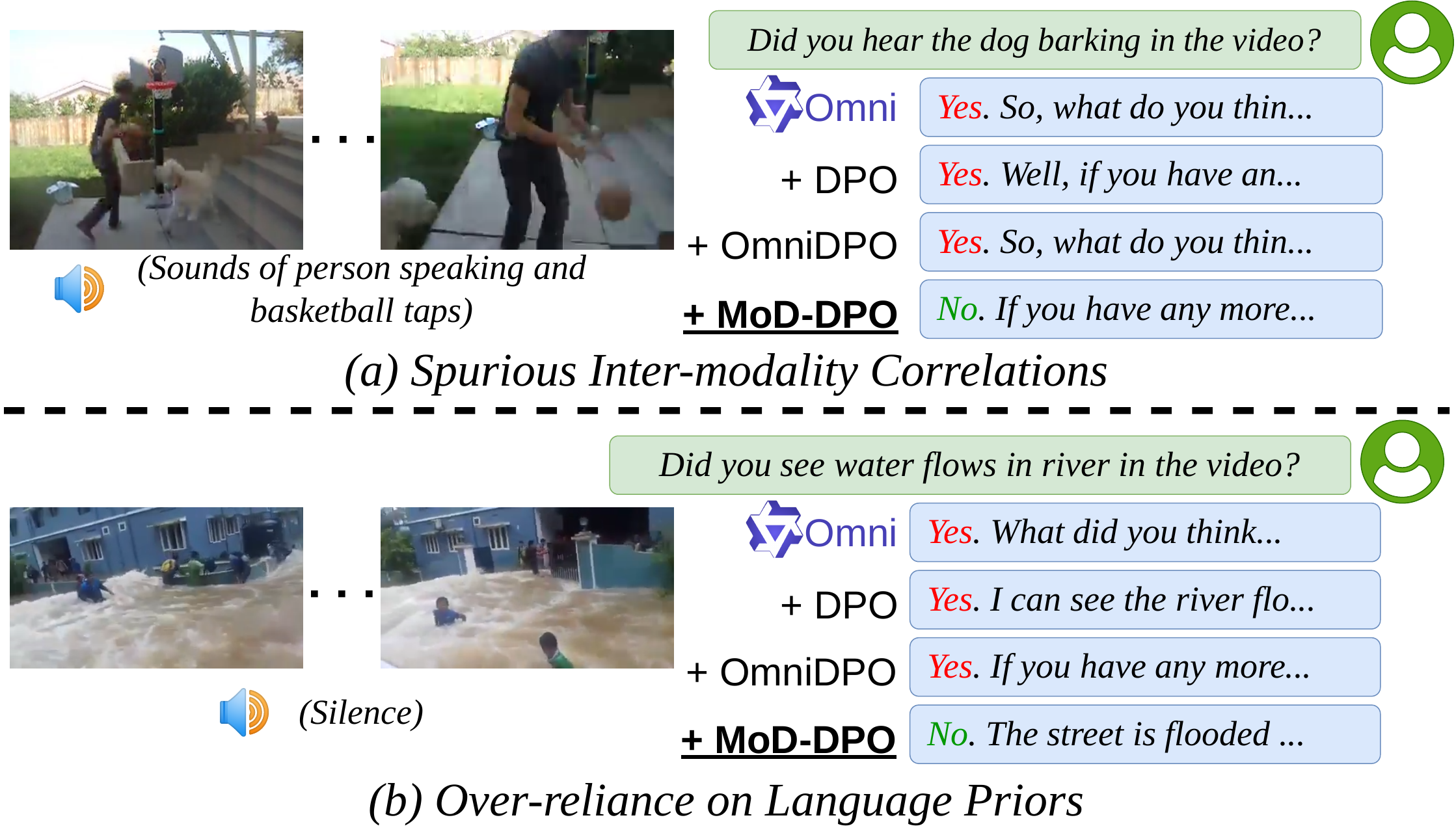}
    \caption{Comparison of the proposed MoD-DPO with other preference optimization baselines for mitigating cross-modal hallucinations arising from spurious inter-modality correlations and over-reliance on language priors.}
    \vspace{-1em}
    \label{fig:qualitative_sample_moddpo}
\end{figure}

Despite impressive capabilities, state-of-the-art models are susceptible to \emph{cross-modal hallucinations} (see \cref{fig:qualitative_sample_moddpo}) when audio and video provide subtle, asynchronous, or weakly correlated evidence \citep{sung-bin2025avhbench, leng2025the_cmmbenchmark}. For example, hearing imaginary sounds from visual cues or perceiving non-existent visual events prompted by audio. Prior analyses \cite{leng2025the_cmmbenchmark} attribute these failures to (i) over-reliance on unimodal priors (especially language), and (ii) spurious inter-modality correlations learned during pretraining and alignment. Existing approaches begin to address these issues by explicitly training models to attend to evidence grounded in the correct modality \cite{sung-bin2025avhbench} or using preference optimization with multimodal preference \cite{chen2025omnidpo}, but they neither \emph{decouple} modality pathways during optimization nor penalize latent language-only shortcuts explicitly. In parallel, decoding-time defenses such as Visual Contrastive Decoding (VCD) \cite{leng2024mitigating_vcd} and its audiovisual extensions \cite{jung2025_avcd} can reduce hallucination without retraining, but they operate post hoc and do not change the model’s internal decision boundaries.

\begin{figure}
    \centering
    \includegraphics[width=\linewidth]{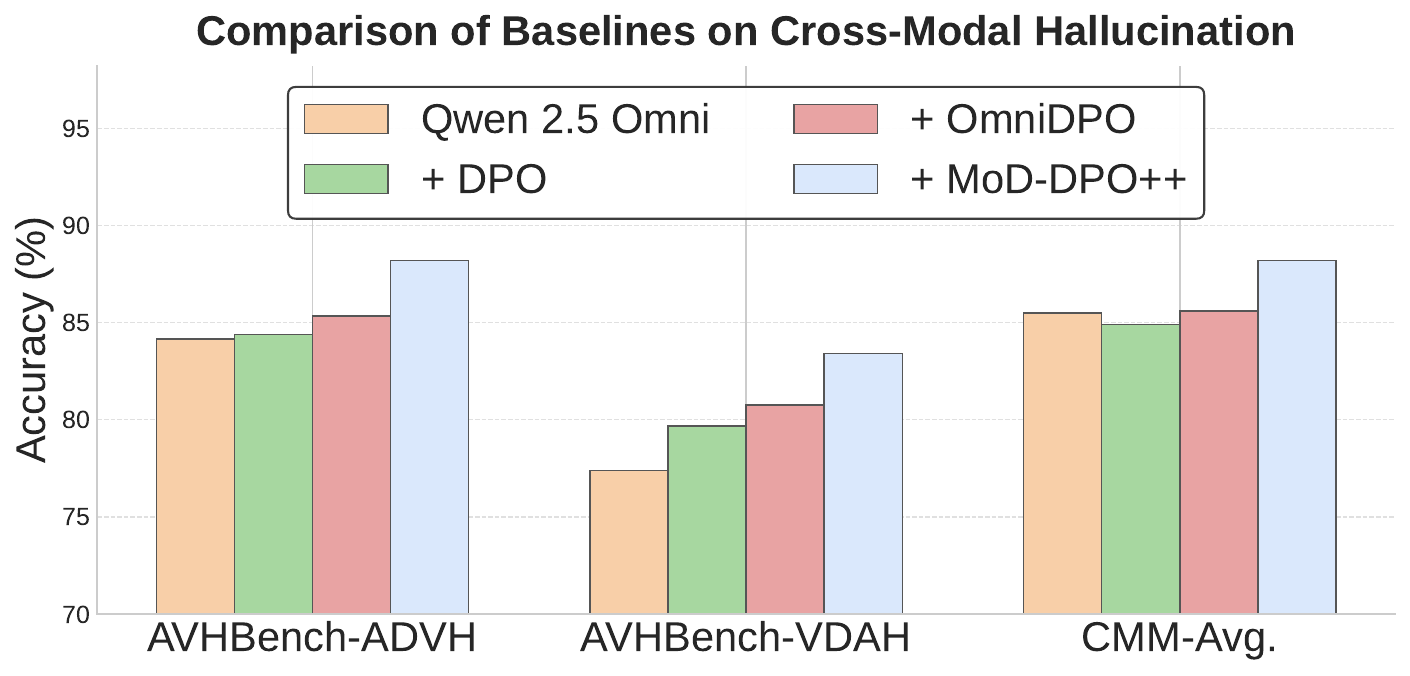}
    \vspace{-1em}
    \caption{Comparison of proposed MoD-DPO++ with other preference optimization baselines on AVHBench \cite{sung-bin2025avhbench} and CMM \cite{leng2025the_cmmbenchmark}. Average accuracy is reported. ADVH: Audio-driven video hallucination, VDAH: Video-driven audio hallucination.}
    \vspace{-2em}
    \label{fig:intro_perf_comparison}
\end{figure}

To remove spurious inter-modality correlations and resulting cross-modal hallucinations, we propose a novel preference optimization technique that \emph{explicitly decouples} modalities during training by enforcing two complementary properties: \emph{invariance} to perturbations of irrelevant modalities and \emph{sensitivity} to the corruption of the relevant modality to generate responses corresponding to prompts related to a modality. Our Modality-Decoupled DPO (MoD-DPO) formalizes these desiderata by adding KL-based regularizers to the original DPO objective \cite{rafailov2023direct_dpo} that (1) keep predictions stable when the unused (irrelevant) modality is corrupted and (2) amplify distributional shifts when the task-relevant modality is corrupted, yielding gradients that nudge the policy toward modality-faithful evidence use. Unlike prior works on multimodal DPO \cite{wang-etal-2024-mdpo,huang2025_vistadpo,chen2025omnidpo}, we derive a closed-form solution to our MoD-DPO objective and optimize using an automatically generated preference dataset.

Since the LLM backbones in omni LLMs are pretrained on large-scale text data, they have strong language priors, which cause over-reliance on language inputs while neglecting audiovisual inputs. To mitigate this issue, we penalize the reward obtained for the closed-form optimal policy during preference optimization \cite{rafailov2023direct_dpo} to ensure that the model responses obtained only using the language inputs are suppressed. 

We evaluate the proposed preference optimization technique on cross-modal hallucination benchmarks -- AVHBench \cite{sung-bin2025avhbench} and Curse of Multi-Modalities (CMM) \cite{leng2025the_cmmbenchmark} -- and report a superior performance of the reference model compared to other post-training techniques as shown in \cref{fig:intro_perf_comparison}. To summarize, the major contributions of our work are as follows.
\begin{itemize}
    \item We propose MoD-DPO, a preference optimization technique to mitigate cross-modal hallucinations in omni LLMs. To support training, we construct a novel preference dataset with over 18.1k automatically generated samples spanning over 10.8k unique videos.
    \item To further reduce the effect of language model priors in omni LLMs, we introduce a language prior debiasing penalty in the preference optimization reward.
    \item We perform experiments on multiple benchmarks to show the efficacy of the proposed approach in mitigating cross-modal hallucinations, outperforming other post-training methods. 
\end{itemize}
\section{Related Works}
\subsection{Multimodal Hallucination}

Large multimodal models often generate content not grounded in the input. Recent works have proposed numerous benchmarks to expose cross-modal failures, in the context of vision-language \cite{li2023evaluatingobjecthallucinationlarge, guan2024hallusionbenchadvanceddiagnosticsuite, chen2024unifiedhallucinationdetectionmultimodal, zhang2025cchallnovelbenchmarkjoint}, audiovisual \cite{sung-bin2025avhbench}, and broad audiovisual-language hallucinations \cite{leng2025the_cmmbenchmark, zhang2025omnievalbenchmarkevaluatingomnimodal}. Another detection technique leverages LLM cross-modal attention patterns to reveal traces of hallucination \cite{zhang2025dhcpdetectinghallucinationscrossmodal}.

To alleviate vision-language hallucinations, two groups of methods have been extensively explored: training-time mitigation and decoding-time defenses. Training-time mitigation focuses on the creation of novel datasets \cite{chen2025perturbollavareducingmultimodalhallucinations, gunjal2024detectingpreventinghallucinationslarge} and the introduction of new learning objectives that attempt to penalize ungrounded generation and reduce prior-induced hallucinations \cite{Yin_2024, huang2024operaalleviatinghallucinationmultimodal, lyu2025alleviatinghallucinationslargevisionlanguage}. Decoding-time defenses include contrastive decoding techniques that reduce inconsistencies between logits and evidence, first observed in vision \cite{leng2024mitigating_vcd} and then in audiovisual settings \cite{jung2025_avcd}. Instruction-contrastive and adaptive focal-contrast methods \cite{chen2024halcobjecthallucinationreduction} further reshape token probabilities at inference time. Visualization-assisted contrastive decoding and retrospective resampling promote self-consistency without retraining \cite{park2024conviscontrastivedecodinghallucination, wu2025generateverifyreducinghallucination}. Additionally, several works extended reinforcement learning from human feedback (RLHF) to multimodal LLMs for hallucination mitigation and human value alignment \cite{yu2024rlhfvtrustworthymllmsbehavior, zhao2025omnialignvenhancedalignmentmllms, ji2024alignanythingtrainingallmodality, zhang2025mmrlhfstepforwardmultimodal}.

\begin{figure*}
    \centering
    \includegraphics[width=\linewidth]{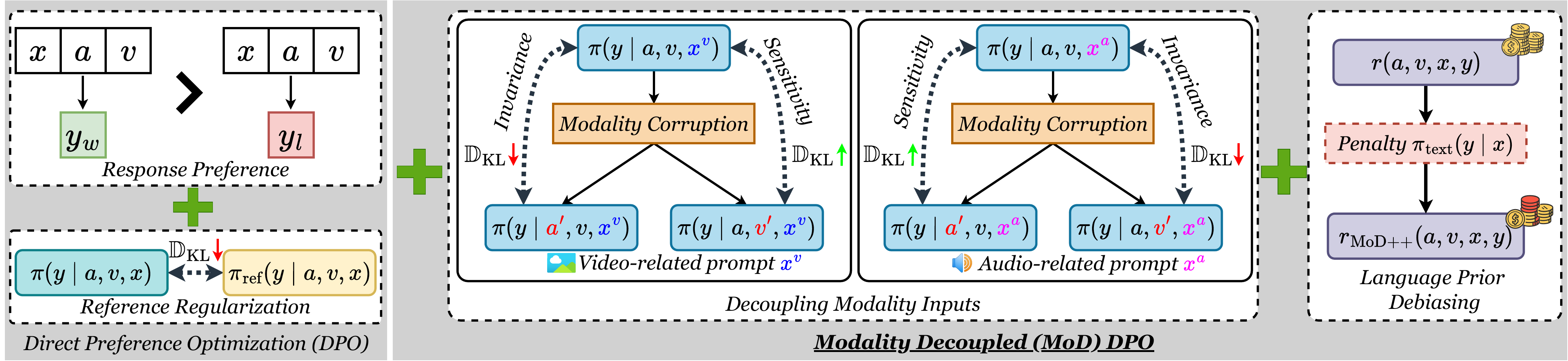}
    \caption{\textbf{Modality Decoupled Preference Optimization.} In addition to the response preference and reference regularization terms in DPO \cite{rafailov2023direct_dpo}, we include additional KL regularization terms to increase model invariance to irrelevant modalities and model sensitivity to relevant modalities. Additionally, we penalize response generation with only text inputs to remove language priors in the model.}
    \label{fig:mod_dpo_main_figure}
\end{figure*}

\subsection{Multimodal Policy Optimization} 

Direct Preference Optimization (DPO) offers a rollout-free alternative to RLHF and has been adapted to mitigate multimodal hallucinations \cite{chen2025omnidpo,huang2025_vistadpo,wang-etal-2024-mdpo}. Instead of costly human labels, many studies derive preference data from strong vision-language referee models, retrieval-augmented contexts, and synthetic preferences produced by reward models and LM-based evaluators \cite{ouali2024clipdpovisionlanguagemodelssource, xing2025realignaligningvisionlanguage, yang2025mitigatinghallucinationslargevisionlanguage, wijaya2024multimodalpreferencedatasynthetic, zhang2024directpreferenceoptimizationvideo}. Several formulations explicitly target hallucinations by skewing preference pairs toward grounded responses or by framing omni-modal mitigation within unified pipelines \cite{fu2024mitigatinghallucinationmultimodallarge, chen2025omnidpo}. However, vanilla multimodal DPO often fails to enforce modality-conditioned preferences, leaving models prone to language-prior shortcuts.

V-DPO \cite{xie2024vdpomitigatinghallucinationlarge} mitigates vision–language hallucinations by injecting video guidance into preference learning through a divergence term that discourages reliance on text-only policies and uses image-contrast pairs to reinforce visual grounding; mDPO \cite{wang-etal-2024-mdpo} strengthens multimodal conditioning via conditional preference terms comparing responses under informative versus degraded images and a reward anchor that prevents down-weighting correct answers; VistaDPO \cite{huang2025_vistadpo} extends preference optimization to videos using hierarchical preferences across instances, time, and regions; and OmniDPO \cite{chen2025omnidpo} generalizes to audio–visual–language settings with text-preference and multimodal-preference pairs that promote attention to visual and auditory evidence. AVEm-DPO \cite{chaubey2026avere} propose to build explicit modality-specific preferences to improve audiovisual emotion understanding. In contrast, our MoD-DPO method is objective-centric and modality-aware, enforcing invariance to perturbations in irrelevant modalities and sensitivity to corruptions in the relevant one.

\label{sec:related_works}
\section{Preliminary}
\label{sec:preliminary}
Direct Preference Optimization (DPO)~\cite{rafailov2023direct_dpo} aligns large language models (LLMs) with human preferences without the need for training a separate reward model.  
Given a reference language model $\pi_{\text{ref}}$, the standard reinforcement learning objective~\cite{jaques2020human_rlhf} for optimizing a policy $\pi_\theta$ is defined as:
\begin{equation}
\small
\begin{aligned}
\max_{\pi_\theta} \;
\mathbb{E}_{x \sim \mathcal{D},\, y \sim \pi_\theta(\cdot \mid x)}
\Big[
r(x,y)
\Big]
- \beta \,
\mathbb{D}_{\text{KL}}\!\left(
\pi_\theta(\cdot \mid x)
\parallel
\pi_{\text{ref}}(\cdot \mid x)
\right),
\end{aligned}
\label{eq:dpo_orig}
\end{equation}
where $r(x,y)$ is the reward assigned to response $y$ for a given prompt $x$, $\beta$ is a hyperparameter and $\mathbb{D}_{\text{KL}}(\cdot \parallel \cdot)$ denotes the Kullback--Leibler divergence~\cite{Kullback1951_kldiv}.  
This objective encourages the model to produce high-reward responses while ensuring that the learned policy $\pi_\theta$ remains close to the reference policy $\pi_{\text{ref}}$. The optimal policy for \cref{eq:dpo_orig} yields the following reward function:
\begin{equation}
\begin{aligned}
r(x, y) = 
\beta \,
\log \frac{\pi(y \mid x)}{\pi_{\text{ref}}(y \mid x)} 
+ \beta \log Z(x),
\end{aligned}
\label{eq:reward_dpo_orig}
\end{equation}
where $Z(x)$ is a partition function independent of both $\pi_\theta$ and $y$. Using the Bradley-Terry \cite{BradleyTerry1952} model to model human preferences results in the following loss function for DPO,
\begin{equation}
\small
\mathcal{L}_{\text{DPO}} = - \mathbb{E} \Bigg[ \log \sigma \Bigg( \beta \Bigg( \log \frac{\pi_\theta(y_w \mid x)}{\pi_{\text{ref}}(y_w \mid x)} - \log \frac{\pi_\theta(y_l \mid x)}{\pi_{\text{ref}}(y_l \mid x)} \Bigg) \Bigg)\Bigg]
\label{eq:dpo_orig_final}
\end{equation}
where the expectation is over a preference dataset $\mathcal{D}^{\text{pref}}_{\text{text}} = \{ (x, y_w, y_l)_i\}$ with $y_w$ as the chosen response and $y_l$ as the rejected response. $\sigma(\cdot)$ denotes the sigmoid function.

\begin{figure*}
    \centering
    \includegraphics[width=0.8\linewidth]{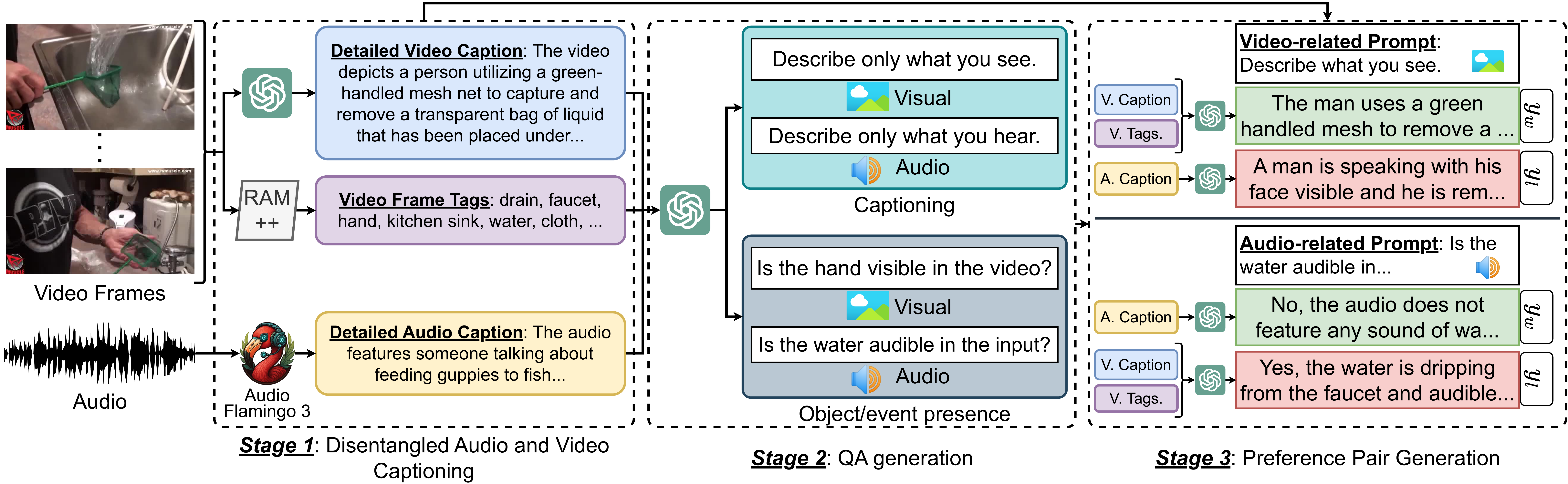}
    \caption{\textbf{Preference Data Generation Pipeline.} We disentangle the audiovisual input to obtain separate audio and visual captions or tags (Stage 1), which are then used to generate QA pairs for Stage 2. Finally, we create preference data for modality-specific questions by constructing chosen responses using relevant modality information and rejected responses using irrelevant modality information.}
    \label{fig:dpo_data_pipeline}
\end{figure*}

\section{Modality Decoupled (MoD) DPO}
\label{sec:method}

Our goal is to mitigate \emph{cross-modal hallucinations} in omni LLMs, which often arise from spurious inter-modal dependencies and over-relying on language model priors \cite{sung-bin2025avhbench,leng2025the_cmmbenchmark}.  
To address this issue, we extend the DPO framework by modifying the original reinforcement learning objective in \cref{eq:dpo_orig} to explicitly decouple modality-specific information in the model’s inputs, thereby reducing unwanted modality interactions.

For an omni LLM that processes both audio and visual inputs, the DPO objective can be generalized as,
\begin{equation}
\small
\begin{aligned}
\max_{\pi_\theta} \;
\mathbb{E} 
\Big[
r(a,v,x,y)
\Big]
- \beta \,
\mathbb{D}_{\text{KL}}\!\left(
\pi_\theta(\cdot \mid a,v,x)
\parallel
\pi_{\text{ref}}(\cdot \mid a,v,x)
\right),
\end{aligned}
\label{eq:dpo_orig_av}
\end{equation}
where $a$ and $v$ denote the audio and visual modalities, respectively, and the expectation is taken over an audiovisual dataset $\mathcal{D}_{\text{av}}$.

\subsection{Decoupling Modality Inputs}
\label{subsec:decoupling_modality_inputs}

We seek to decouple modality information by enforcing two key properties:
\begin{enumerate}[label=(\roman*)]
    \item \textbf{Invariance:} The output distribution should remain stable when the \emph{prompt-irrelevant} modality is corrupted, making the model response agnostic to changes in irrelevant input information.
    \item \textbf{Sensitivity:} The output distribution should shift appropriately when the \emph{prompt-relevant} modality is corrupted, making the model response more sensitive to important input information.
\end{enumerate}

Specifically, let us assume that the input prompt $x^v$ is related to the visual modality. Then our objective in \cref{eq:dpo_orig_av} changes to the following,
\begin{equation}
\begin{aligned}
\max_{\pi_\theta} &\; 
\mathbb{E}_{(a,v,x^v)\sim\mathcal{D},y\sim \pi_\theta(\cdot\mid a,v,x^v)}\left[
r(a,v,x^v,y)\right] \\[-4pt]
&- \beta \mathbb{D}_{\text{KL}}(\pi_\theta(\cdot \mid a,v,x^v) \parallel \pi_{\text{ref}}(\cdot \mid a,v,x^v)) \\[-4pt]
&\colorbox{green!25}{$
-\beta_{\text{inv}}\,\mathbb{D}_{\text{KL}}(\pi_\theta(\cdot \mid a,v,x^v) \parallel \pi_\theta(\cdot \mid a',v,x^v))
$} \\[-4pt]
&\colorbox{blue!25}{$
+\beta_{\text{sens}}\,\mathbb{D}_{\text{KL}}(\pi_\theta(\cdot \mid a,v,x^v) \parallel \pi_\theta(\cdot \mid a,v',x^v))
$},
\end{aligned}
\label{eq:dpo_mod}
\end{equation}
where $a'$ and $v'$ denote corrupted versions of the audio and visual inputs, respectively. The green-highlighted term enforces \emph{invariance} to corruption in the irrelevant modality (audio), while the blue-highlighted term enforces \emph{sensitivity} to corruption in the relevant modality (visual).  
The hyperparameters $\beta_{\text{inv}}$ and $\beta_{\text{sens}}$ control the respective strengths of these regularization effects. \cref{fig:mod_dpo_main_figure} shows the comparison of the proposed approach with vanilla DPO.

The optimal policy $\pi^*$ for \cref{eq:dpo_mod} can be obtained using the method of Lagrange multipliers.  
However, unlike the standard DPO objective in \cref{eq:dpo_orig}, this modified formulation introduces additional dependencies on $\pi_\theta(y \mid a', v, x)$ and $\pi_\theta(y \mid a, v', x)$, which complicate the optimization process.  
To address this challenge, we assume these terms remain constant within each optimization step and treat them as fixed target distributions when computing the additional KL divergence terms in \cref{eq:dpo_mod}.  

Under this assumption, we can take the partial derivative of \cref{eq:dpo_mod} with respect to $\pi_\theta(y \mid a, v, x^v)$ and solve for the stationary point, yielding the following optimal policy,
\begin{equation}
\small
\begin{aligned}
\pi^\star_\theta(y \mid a, v, x^v)
&\propto\;
\exp\!\big(r(a, v, x^v, y) / \tau \big)\,
\pi_{\text{ref}}(y \mid a, v, x^v)^{\beta / \tau} \\
&
\colorbox{green!25}{$\pi'_\theta(y \mid a', v, x^v)^{\beta_{\text{inv}} / \tau}$}\,
\colorbox{blue!25}{$\pi'_\theta(y \mid a, v', x^v)^{-\beta_{\text{sens}} / \tau}$},
\end{aligned}
\label{eq:dpo_mod_opt_policy}
\end{equation}
where $\tau = \beta + \beta_{\text{inv}} + \beta_{\text{sens}}$, and $\pi'_\theta$ denotes the fixed target distributions corresponding to corrupted audiovisual inputs. We detail the above proof in \appref{app:opt_policy_proof}. 

Inspecting \cref{eq:dpo_mod_opt_policy}, we observe that the optimal policy increases the likelihood of correct responses when the irrelevant modality is corrupted (invariance) while penalizing the likelihood when the relevant modality is corrupted (sensitivity). This balance encourages the model to focus on modality-relevant cues while suppressing spurious cross-modal dependencies. 

We can find the optimal reward function for \cref{eq:dpo_mod_opt_policy} as the following,
\begin{equation}
\small
\begin{aligned}
r_{\text{MoD}}(a,v,&x^{v},y)
=\tau\,\log \pi_\theta(y\mid a,v,x^v)
-\beta\,\log \pi_{\text{ref}}(y\mid a,v,x^v) \\
&\colorbox{green!25}{$-\beta_{\text{inv}}\,\log \pi'_\theta(y\mid a',v,x^v)$}
\colorbox{blue!25}{$+\beta_{\text{sens}}\,\log \pi'_\theta(y\mid a,v',x^v)$} \\
&+ W(a,v,x), \label{eq:reward_dpo_mod}
\end{aligned}
\end{equation}
where $W(a,v,x)$ is obtained by normalizing \cref{eq:dpo_mod_opt_policy}. Please refer to the \appref{app:reward_proof} \& \ref{app:bt_loss_proof} for more details. We can substitute this reward function into the Bradley-Terry model, similar to \cref{eq:dpo_orig_final} to obtain the loss function for modality decoupled preference optimization as follows,
\begin{equation}
\footnotesize
\begin{aligned}
\mathcal{L}_{\text{MoD}}^{v}
&= -\mathbb{E}\Bigg[
\log \sigma\!\Bigg( \tau
\log \frac{\pi_{\theta}(y_w \mid a,v,x^v)}{\pi_{\theta}(y_l \mid a,v,x^v)}
-\beta \log \frac{\pi_{\text{ref}}(y_w \mid a,v,x^v)}{\pi_{\text{ref}}(y_l \mid a,v,x^v)} \\[-3pt]
&\quad
\colorbox{green!25}{$-\beta_{\text{inv}} \log \frac{\pi'_{\theta}(y_w \mid a',v,x^v)}{\pi'_{\theta}(y_l \mid a',v,x^v)}$}
\colorbox{blue!25}{$+\beta_{\text{sens}} \log \frac{\pi'_{\theta}(y_w \mid a,v',x^v)}{\pi'_{\theta}(y_l \mid a,v',x^v)}$}
\Bigg)
\Bigg],
\label{eq:dpo_mod_vision_final}
\end{aligned}
\end{equation}
where $\tau = \beta + \beta_{\text{inv}} - \beta_{\text{sens}}$.
Note that this objective is only for prompts related to the visual modality. We can write a similar form for prompts $x^a$ related to the audio modality as $\mathcal{L}_{\text{MoD}}^{a}$ (\appref{app:audio_objective}). The final loss function for decoupling modality inputs is as follows,
\begin{equation}
    \mathcal{L}_{\text{MoD}} = \mathcal{L}_{\text{MoD}}^{v} + \mathcal{L}_{\text{MoD}}^{a}
\label{eq:loss_dpo_mod_decoupled}
\end{equation}
\vspace{-2em}
\paragraph{Practical considerations.} To ensure that $\pi'_\theta$ (in \cref{eq:dpo_mod_opt_policy,eq:reward_dpo_mod,eq:dpo_mod_vision_final}) is a fixed target during training for an iteration, we stop gradient computation and accumulation when passing the corrupt versions of audiovisual inputs ($a'$, $v'$) to the model being trained. Moreover, during training, we alternate between training a batch of vision-related prompts $x^v$ and audio-related prompts $x^a$ rather than having a mixed batch. Additionally, we set $\beta_{\text{inv}}$ lower than $\beta_{\text{sens}}$, as cross-modal information is sometimes beneficial even if the prompts are specific to a particular modality. For example, the presence of a visual entity (such as a dog) can be confirmed by its sound (barking) in the audio. Please see \cref{subsubsec:strength_hyperparam} (and \appref{app_subsec:hyperparam_tuning}) for empirical analysis.
\vspace{-1em}
\paragraph{What about audiovisual inputs?} We only perform DPO training on queries strongly linked to one of the modalities. We can modify the objective in \cref{eq:dpo_mod} to get $\mathcal{L}^{av}_{\text{MoD}}$ for audiovisual tasks (like audiovisual captioning) as shown in \appref{app:audiovisual_dpo_mod_objective}. However, our empirical results show that including such an objective during DPO training results in insignificant performance gains for eliminating cross-modal hallucinations (\cref{para:joint_audiovisual_training,tab:match_mismatch_ablation}).


\begin{figure}[t]
    \centering
    \begin{subfigure}{0.58\linewidth}
        \centering
        \includegraphics[width=\linewidth]{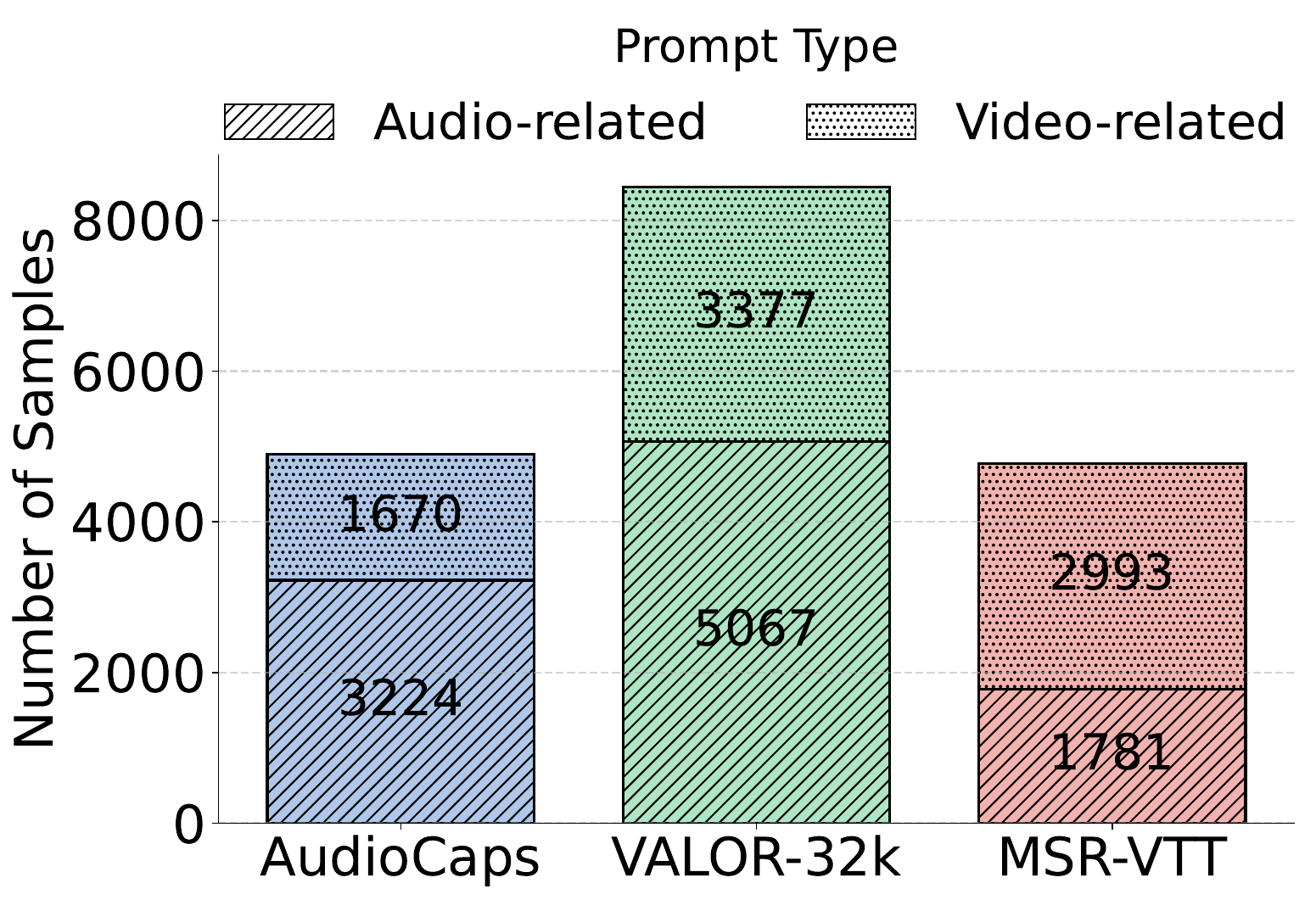}
    \end{subfigure}
    \hfill
    \begin{subfigure}{0.38\linewidth}
        \centering
        \includegraphics[width=\linewidth]{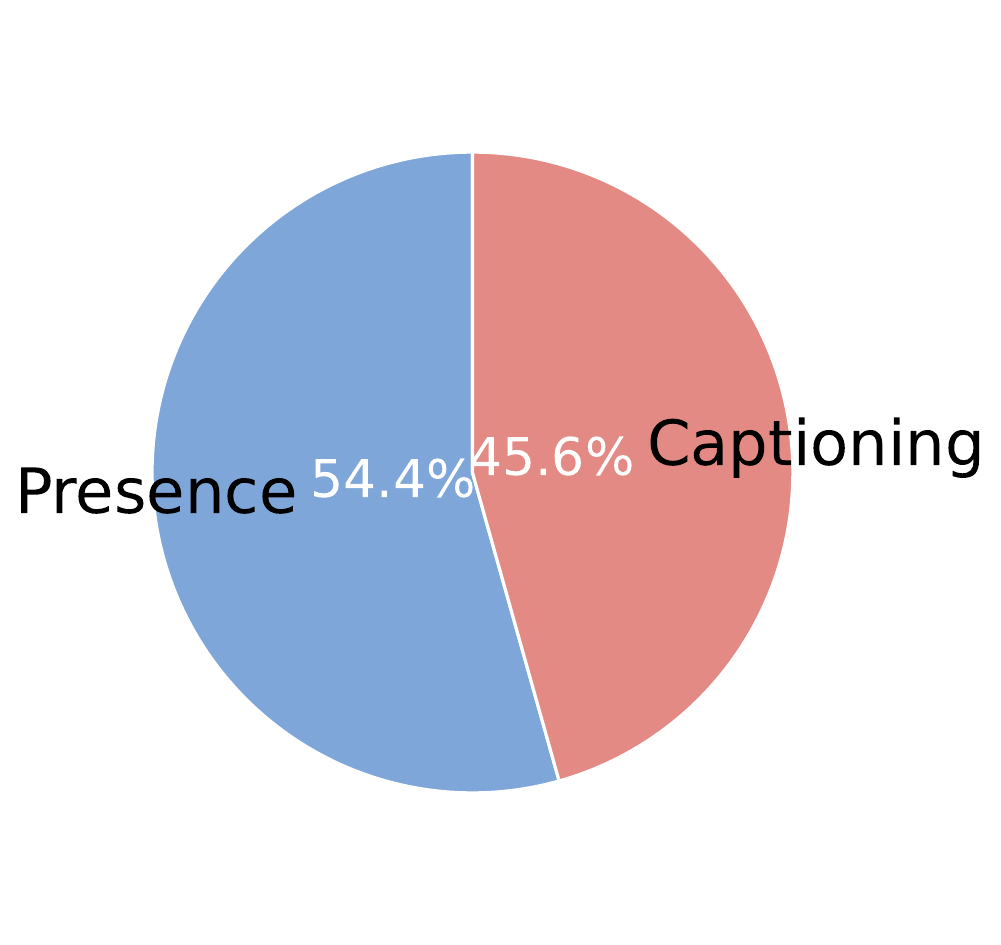}
    \end{subfigure}
    \caption{\textbf{Preference Data Statistics.} \emph{(Left)} Number of preference samples generated from different source datasets. \emph{(Right)} Composition of samples belonging to different tasks. }
    \label{fig:dpo_data_stats}
\end{figure}

\subsection{Language-Prior Debiasing (LPD)}
\label{subsec:language_prior_debiasing}
To ensure that the language priors in the omni LLM do not solely control the outputs \cite{leng2025the_cmmbenchmark} (e.g., \cref{fig:qualitative_sample_moddpo} - bottom), we introduce a language prior debiasing penalty to the preference optimization reward $r(a,v,x^v,y)$ in \cref{eq:reward_dpo_mod} to reduce the log-probability of the chosen response given just the text inputs $x^v$ (\cref{fig:mod_dpo_main_figure}). More concretely, our LPD penalty for a given input-response pair is as follows,
\begin{equation}
\label{eq:lpd_penalty}
r_{\text{LPD}} (x^v,y) = \colorbox{red!25}{$- \log \pi_{\text{text}} (y \mid x^v)$},
\end{equation}
where $\pi_{\text{text}}$ is a language model which can take text-only inputs. Note that $\pi_{\text{text}}$ is a frozen language model which captures language priors, and theoretically it can be any strong language model. However, we choose $\pi_{\text{text}}$ as the reference model $\pi_{\text{ref}}$ for efficiency. The total reward in \cref{eq:reward_dpo_mod} using this penalty becomes,
\begin{equation}
\small
\begin{aligned}
r_{\text{MoD++}}(a,v,x^{v},y)
=  r_{\text{MoD}}(a,v,x^{v},y) + \gamma_{\text{LPD}} \; r_{\text{LPD}} (x^v,y),
\label{eq:reward_dpo_mod_pp}
\end{aligned}
\end{equation}
where $\gamma_{\text{LPD}}$ is a hyperparameter to control language prior debiasing. The final objective (\textbf{MoD-DPO++}) with additional language prior debiasing becomes the following,
\begin{equation}
\mathcal{L}_{\text{MoD++}} = \mathcal{L}_{\text{MoD}} \colorbox{red!25}{$- \gamma_{\text{LPD}} \; \log \frac{\pi_{\text{ref}} (y_w \mid x)}{\pi_{\text{ref}} (y_l \mid x)}$},
\label{eq:loss_dpo_mod_pp}
\end{equation}
where $x$ is a combined representation to denote prompts related to audio ($x^a$) and video ($x^v$).

\label{para:complecity_analysis}
\paragraph{Complexity Analysis.} The proposed MoD-DPO++ framework introduces additional computational overhead per preference tuning iteration compared to standard DPO. Specifically, our objective requires additional forward passes for corrupted audio and visual inputs to enforce modality invariance and sensitivity. Moreover, the language prior debiasing (LPD) term involves an extra forward pass using text-only inputs. However, since none of the additional forward passes require gradient computation, additional gradient computation compared to Vanilla DPO is minimal. To maintain comparable training budgets, we therefore train MoD-DPO++ for only one quarter of the total epochs used by baseline methods. Please refer to \appref{app_subsec:training_compute_details} for detailed analysis.

\begin{table*}[]
\centering
\caption{Comparison of the proposed approach with other DPO and omni LLM methods on AVHBench \cite{sung-bin2025avhbench}.}
\vspace{-1em}
\resizebox{\textwidth}{!}{%
\begin{tabular}{l|cccc|cccc|cccc}
\hline\hline
\rowcolor[HTML]{C0C0C0} 
\multicolumn{1}{c|}{\cellcolor[HTML]{C0C0C0}} & \multicolumn{4}{c|}{\cellcolor[HTML]{C0C0C0}\textbf{Audio-driven Video Hallucination}} & \multicolumn{4}{c|}{\cellcolor[HTML]{C0C0C0}\textbf{Video-driven Audio Hallucination}} & \multicolumn{4}{c}{\cellcolor[HTML]{C0C0C0}\textbf{Audiovisual Matching}} \\ \cline{2-13} 
\rowcolor[HTML]{C0C0C0} 
\multicolumn{1}{c|}{\multirow{-2}{*}{\cellcolor[HTML]{C0C0C0}\textbf{Method}}} & \textbf{Acc.} & \textbf{Pre.} & \textbf{Rec.} & \textbf{F1} & \textbf{Acc.} & \textbf{Pre.} & \textbf{Rec.} & \textbf{F1} & \textbf{Acc.} & \textbf{Pre.} & \textbf{Rec.} & \textbf{F1} \\ \hline
VideoLLaMA 2 \cite{cheng2024videollama2} & 79.23 & 81.51 & 76.94 & 79.16  & 75.07 & 85.15 & 64.98 & 73.71  & 52.93 & 92.11 & 13.75 & 23.93  \\
VITA-1.5 \cite{fu2025vita1.5} & 67.17 & 78.17 & 56.17 & 65.36 & 54.01 & 45.85 & 62.18 & 52.78 & 46.85 & 66.31 & 27.40 & 38.78 \\
OmniVinci \cite{omnivinci2025} & 61.36 & 57.92 & 64.79 & 61.16 & 58.56 & 84.71 & 32.40 & 46.88 & 54.32 & 87.42 & 21.21 & 34.14 \\
Qwen 3 Omni \cite{xu2025qwen3omni} & 83.54 & 85.92 & 81.17 & 83.47 & 76.46 & 94.94 & 57.99 & 72.00 & 58.52 & 97.76 & 19.14 & 32.02 \\ \hline
Qwen 2.5 Omni \cite{xu2025qwen25omnitechnicalreport} & 84.15 & 76.76 & 91.55 & 83.51 & 77.38 & \textbf{94.93} & 59.83 & 73.39 & 54.69 & 9.81 & \textbf{99.57} & 17.85 \\
+ DPO \cite{rafailov2023direct_dpo} & 84.39 & 75.44 & 93.30 & 83.42 & 79.68 & 93.53 & 65.85 & 77.28 & 59.32 & 19.76 & 98.93 & 32.94 \\
+ OmniDPO \cite{chen2025omnidpo} & 85.34 & 75.61 & \textbf{95.06} & 84.23 & 80.77 & 86.36 & \textbf{75.19} & 80.39 & 61.50 & 23.90 & 99.14 & 38.52 \\
\rowcolor[HTML]{DAE8FC} 
+ MoD-DPO & 87.66 & 85.53 & 89.78 & 87.61 & 82.48 & 93.61 & 71.35 & 80.98 & 69.07 & 42.09 & 96.04 & 58.53\\
\rowcolor[HTML]{DAE8FC} 
+ MoD-DPO++ & \textbf{88.19} & \textbf{86.35} & 90.02 & \textbf{88.15} & \textbf{83.40} & \textbf{93.82} & 72.98 & \textbf{82.10} & \textbf{69.68} & \textbf{43.32} & 96.04 & \textbf{59.71} \\ \hline
MiniCPM-O 2.6 \cite{yao2024minicpm} & 83.36 & 85.56 & 81.16 & 83.30 & 74.54 & 82.27 & 66.81 & 73.74 & 54.26 & 56.72 & 51.81 & 54.15 \\
+ DPO \cite{rafailov2023direct_dpo} & 82.91 & 84.56 & 81.26 & 82.88 & 78.86 & \textbf{82.82} & 68.89 & 75.22 & 54.56 & 55.91 & 53.20 & 54.52 \\
+ OmniDPO \cite{chen2025omnidpo} & 84.96 & 86.18 & 83.75 & 84.95 & 75.39 & 80.56 & 70.22 & 75.04 & 56.86 & 57.82 & 55.90 & 56.84 \\
\rowcolor[HTML]{DAE8FC} 
+ MoD-DPO & {87.08} & \textbf{89.28} & {84.87} & {87.02} & {79.00} & 82.18 & {75.81} & {78.87} & 60.57 & \textbf{62.16} & {58.98} & {60.53} \\
\rowcolor[HTML]{DAE8FC} 
+ MoD-DPO++ & \textbf{87.26} & 88.89 & \textbf{85.63} & \textbf{87.23} & \textbf{79.49} & 82.39 & \textbf{76.58} & \textbf{79.38} & \textbf{60.66} & 61.87 & \textbf{59.45} & \textbf{60.64}  \\
\hline\hline
\end{tabular}%
}
\label{tab:avhbench_main_res}
\vspace{-0.5em}
\end{table*}

\subsection{Training Dataset}
\label{subsec:dpo_preference_dataset}
\cref{fig:dpo_data_pipeline} shows the preference data generation pipeline for performing the proposed preference optimization, comprising three stages. We ensure that the audiovisual information is decoupled during data generation to prevent the model from learning inter-modality spurious correlations. We summarize the stages of data generation in the following paragraphs with additional details present in \appref{app_subsec:detailed_data_stages}.

\noindent\textbf{Stage 1.} In the first stage,  audio and visual inputs are separated and automatically annotated. For videos, we extract the visual captions through GPT-4o \cite{hurst2024gpt4o} and the visual tags through RAM++ \cite{zhang2024recognize_ram,huang2025_rampp}. For audio, we extract the audio captions through AudioFlamingo 3 \cite{ghosh2025audioflamingo3_af3}. We source the original audiovisual files from MSR-VTT \cite{xu2016msrvtt}, VALOR32K (val) \cite{chen2023valor32k} and AudioCaps (val) \cite{kim-etal-2019-audiocaps}.  In addition to the original audiovisual inputs with coupled audio and visual information, in most cases, we create mismatched audiovisual contexts using audio and video segments from different files, leading to improved performance (\cref{para:importance_of_mismatched_context}). 

\noindent\textbf{Stage 2.} We automatically generate QA pairs using GPT-4o \cite{hurst2024gpt4o} for two tasks -- audio/visual captioning and audio/visual object/event presence. These tasks are fundamental for audiovisual understanding, and as our results show, training on these tasks results in improvement in general audio and video tasks (\cref{subsec:results_general_benchmarks}) in addition to cross-modal hallucination benchmarks. Audio and visual information extracted in Stage 1 is provided to GPT-4o along with QA generation instructions.

\noindent \textbf{Stage 3.} Instead of generating rejected responses $y_l$ which are completely irrelevant to an input prompt, we generate hard-negative rejected responses which include spurious information from the information extracted in the irrelevant modality. For example, to generate the rejected response for a prompt related to the visual modality, we create a rejected response by including information from the audio modality. 

\noindent\textbf{Data Statistics.} We generate a total of 18,112 preference data samples over 10,854 unique videos using the above pipeline. \cref{fig:dpo_data_stats} shows the statistics of the preference dataset. Preference data samples are present in \appref{app_subsec:pref_data_samples}.

\section{Experimental Details}

\subsection{Evaluation Benchmarks and Metrics}

\noindent \textbf{Benchmarks.} We evaluate our post-trained model for cross-modal hallucinations using AVHBench \cite{sung-bin2025avhbench} and Curse of Multi-Modaliites (CMM) Benchmark \cite{leng2025the_cmmbenchmark}. AVHBench tests cross-modal hallucination and audiovisual context matching with around 5k samples over 2k unique videos. CMM consists of tasks to test over-reliance on unimodal priors and spurious inter-modal correlations with 2.4k samples over 1.2k unique videos. For testing the general audio/visual capabilities of the model after DPO, we use DailyOmni \cite{zhou2025dailyomni} for joint audiovisual understanding, MVBench \cite{li2024mvbench} for visual understanding, and MMAU \cite{sakshi2025mmau} for audio understanding.

\noindent \textbf{Evaluation Metrics.} We report the accuracy, precision, recall, and F1 over the different tasks in AVHBench. 
For CMM \cite{leng2025the_cmmbenchmark}, we report the perception accuracy (pa) and hallucination resistance (hr) over all the tasks. For general datasets, we report the average accuracy over different tasks of the dataset. Refer \appref{app_subsec:eval_protocol} for more details.


\subsection{Implementation Details}
We modify the LLaMAFactory \cite{zheng2024llamafactory} codebase to implement the proposed method and to report results on the preference optimization baselines. We perform our experiments using two reference models -- Qwen 2.5 Omni \cite{xu2025qwen25omnitechnicalreport} (7B params) and MiniCPM-O 2.6 \cite{yao2024minicpm} (8B params). We train the proposed MoD-DPO with a learning rate of $3e^{-7}$ for a single epoch on the preference data using a batch size of 1 per GPU on 8 H100 GPUs. As discussed in \cref{para:complecity_analysis}, for a fair comparison with MoD-DPO++, we train the baseline preference optimization techniques for four epochs. We choose $\beta=0.1$ consistent with other DPO baselines \cite{huang2025_vistadpo,wang-etal-2024-mdpo}, where as $\beta_{\text{sens}}=0.05$, $\beta_{\text{inv}}=0.02$, and $\gamma_{\text{LPD}}=0.05$ based on hyperparameter tuning (\cref{subsubsec:strength_hyperparam}).

\begin{table*}[]
\centering
\caption{Comparison of the proposed approach with other DPO and omni LLM methods on Curse of Multi-Modalities \cite{leng2025the_cmmbenchmark}.}
\vspace{-1em}
\label{tab:results_main_cmm}
\resizebox{0.9\textwidth}{!}{%
\begin{tabular}{l|cc|cc|cc|cc|cc|cc|cc}
\hline\hline 
\rowcolor[HTML]{C0C0C0} 
\multicolumn{1}{c|}{\cellcolor[HTML]{C0C0C0}} & \multicolumn{6}{c|}{\cellcolor[HTML]{C0C0C0}\textbf{Spurious Inter-modality Correlation}} & \multicolumn{6}{c|}{\cellcolor[HTML]{C0C0C0}\textbf{Uni-modality Over-reliance}} & \multicolumn{2}{c}{\cellcolor[HTML]{C0C0C0}\textbf{Overall}} \\ \cline{2-15} 
\rowcolor[HTML]{C0C0C0} 
\multicolumn{1}{c|}{\cellcolor[HTML]{C0C0C0}} & \multicolumn{2}{c|}{\cellcolor[HTML]{C0C0C0}\textbf{VL}} & \multicolumn{2}{c|}{\cellcolor[HTML]{C0C0C0}\textbf{AL}} & \multicolumn{2}{c|}{\cellcolor[HTML]{C0C0C0}\textbf{VAL}} & \multicolumn{2}{c|}{\cellcolor[HTML]{C0C0C0}\textbf{Visual Dom.}} & \multicolumn{2}{c|}{\cellcolor[HTML]{C0C0C0}\textbf{Audio Dom.}} & \multicolumn{2}{c|}{\cellcolor[HTML]{C0C0C0}\textbf{Lang. Dom.}} & \cellcolor[HTML]{C0C0C0} & \cellcolor[HTML]{C0C0C0} \\
\rowcolor[HTML]{C0C0C0} 
\multicolumn{1}{c|}{\multirow{-3}{*}{\cellcolor[HTML]{C0C0C0}\textbf{Method}}} & \textbf{pa} & \multicolumn{1}{c|}{\cellcolor[HTML]{C0C0C0}\textbf{hr}} & \textbf{pa} & \multicolumn{1}{c|}{\cellcolor[HTML]{C0C0C0}\textbf{hr}} & \textbf{pa} & \textbf{hr} & \textbf{pa} & \multicolumn{1}{c|}{\cellcolor[HTML]{C0C0C0}\textbf{hr}} & \textbf{pa} & \multicolumn{1}{c|}{\cellcolor[HTML]{C0C0C0}\textbf{hr}} & \textbf{pa} & \textbf{hr} & \multirow{-2}{*}{\cellcolor[HTML]{C0C0C0}\textbf{pa}} & \multirow{-2}{*}{\cellcolor[HTML]{C0C0C0}\textbf{hr}} \\ \hline
VideoLLaMA 2 \cite{cheng2024videollama2} & 75.0 & 86.0 & 77.5 & 94.0 & 78.0 & 98.0 & 62.0 & 75.5 & 80.0 & 90.0 & 57.5 & 43.0 & 71.7 & 81.1 \\
VITA 1.5 \cite{fu2025vita1.5} & 92.5 & 91.5 & 42.5 & 30.5 & 77.7 & 72.9 & 57.0 & 24.4 & 77.4 & 67.7 & 86.0 & 55.5 & 72.2 & 57.1 \\
OmniVinci \cite{omnivinci2025} & 88.5 & 81.0 & 74.5 & 90.0 & 98.0 & 72.0 & 91.0 & 63.5 & 96.0 & 35.0 & 87.0 & 75.0 & 89.2 & 69.4 \\
Qwen 3 Omni \cite{xu2025qwen3omni} & 96.5 & 77.5 & 93.0 & 88.0 & 97.5 & 94.0 & 95.5 & 60.0 & 94.5 & 67.0 & 93.0 & 65.0 & 95.0 & 75.3 \\ \hline
Qwen 2.5 Omni \cite{xu2025qwen25omnitechnicalreport} & 88.0 & \multicolumn{1}{c|}{98.0} & 89.9 & \multicolumn{1}{c|}{84.4} & 93.0 & 93.5 & 87.8 & \multicolumn{1}{c|}{67.7} & 85.3 & \multicolumn{1}{c|}{80.7} & {74.2} & 83.2 & 86.4 & 84.6 \\
+ DPO \cite{rafailov2023direct_dpo} & 86.0 & \multicolumn{1}{c|}{98.5} & 89.5 & \multicolumn{1}{c|}{\textbf{86.5}} & 92.4 & 94.4 & 86.7 & \multicolumn{1}{c|}{63.5} & 84.3 & \multicolumn{1}{c|}{80.4} & 72.5 & \textbf{84.1} & 85.2 & 84.6 \\
+ OmniDPO \cite{chen2025omnidpo} & 89.0 & \multicolumn{1}{c|}{98.0} & 89.5 & \multicolumn{1}{c|}{83.5} & 94.0 & 93.0 & 88.8 & \multicolumn{1}{c|}{69.9} & 86.1 & \multicolumn{1}{c|}{81.3} & 72.2 & 82.6 & 86.6 & 84.7 \\
\rowcolor[HTML]{DAE8FC} 
+ MoD-DPO & {91.5} & \multicolumn{1}{c|}{\textbf{98.5}} & {92.3} & \multicolumn{1}{c|}{83.2} & \textbf{94.5} & {94.5} & \textbf{92.3} & \multicolumn{1}{c|}{74.2} & {88.4} & \multicolumn{1}{c|}{83.8} & 73.9 & 83.1 & {88.8} & {86.2} \\
\rowcolor[HTML]{DAE8FC}
+ MoD-DPO++ & \textbf{92.5} & \multicolumn{1}{c|}{\textbf{98.5}} & \textbf{92.7} & \multicolumn{1}{c|}{85.8} & \textbf{94.5} & \textbf{95.5} & {91.3} & \multicolumn{1}{c|}{\textbf{75.1}} & \textbf{89.1} & \multicolumn{1}{c|}{\textbf{84.3}} & \textbf{74.9} & 83.8 & \textbf{89.2} & \textbf{87.2} \\
\hline
MiniCPM-O 2.6 \cite{yao2024minicpm} & 87.0 & 94.0 & 90.5 & 74.0 & 87.0 & 84.0 & 79.5 & 80.5 & 87.4 & 76.4 & 82.0 & 73.7 & 85.6 & 80.4 \\
+ DPO \cite{rafailov2023direct_dpo} & 87.0 & 93.0 & 89.5 & 74.5 & 86.0 & 84.5 & 79.5 & 79.5 & 87.8 & 76.0 & 81.5 & {74.3} & 85.2 & 80.3 \\
+ OmniDPO \cite{chen2025omnidpo} & 88.0 & 95.0 & 91.0 & {76.5} & 88.5 & 84.0 & 79.1 & 78.8 & 88.4 & 77.8 & {83.2} & 71.6 & 86.4 & 80.6 \\
\rowcolor[HTML]{DAE8FC}
+ MoD-DPO & \textbf{90.5} & {95.5} & {92.0} & 76.0 & \textbf{91.5} & {86.0} & {82.5} & {84.8} & \textbf{89.0} & {78.9} & 82.4 & 73.7 & {88.0} & {82.5} \\
\rowcolor[HTML]{DAE8FC}
+ MoD-DPO++ & {90.0} & \textbf{96.5} & \textbf{92.5} & \textbf{77.5} & \textbf{91.5} & \textbf{87.0} & \textbf{83.5} & \textbf{85.6} & {88.4} & \textbf{79.6} & \textbf{83.8} & \textbf{75.1} & \textbf{88.3} & \textbf{83.6} \\
\hline\hline
\end{tabular}%
\vspace{-2em}
}
\end{table*}

\section{Results}

\subsection{Cross-Modal Hallucination Benchmarks}

\cref{tab:avhbench_main_res} shows the performance of the proposed approach on AVHBench \cite{sung-bin2025avhbench} with comparison to several baselines. We can clearly see that MoD-DPO and MoD-DPO++ outperform all other baselines for overall accuracy and F1, showing a balanced performance. MoD-DPO results in gains of up to 27\% accuracy relative to the reference models on the audiovisual matching task. On the CMM \cite{leng2025the_cmmbenchmark} Benchmark in \cref{tab:results_main_cmm}, we can see similar trends with around 3-4 \% overall performance gain over the reference models. Note that for the language dominance task, the improvement of MoD-DPO++ over MoD-DPO is significantly higher than in other tasks, thus showing the efficacy of LPD (\cref{subsec:language_prior_debiasing}) in reducing language priors.

\begin{table}[]
\centering
\caption{Ablation study shows performance improvement of Qwen 2.5 Omni \cite{xu2025qwen25omnitechnicalreport} using different components of the proposed MoD-DPO++. LPD: Language Prior Debiasing.}
\vspace{-1em}
\label{tab:ablation_study}
\resizebox{\columnwidth}{!}{%
\begin{tabular}{ccc|ccc|cc}
\hline \hline
\rowcolor[HTML]{C0C0C0} 
\cellcolor[HTML]{C0C0C0} & \cellcolor[HTML]{C0C0C0} & \cellcolor[HTML]{C0C0C0} & \multicolumn{3}{c|}{\cellcolor[HTML]{C0C0C0}\textbf{AVHBench}\cite{sung-bin2025avhbench}} & \multicolumn{2}{c}{\cellcolor[HTML]{C0C0C0}\textbf{CMM}\cite{leng2025the_cmmbenchmark}} \\ \cline{4-8} 
\rowcolor[HTML]{C0C0C0} 
\multirow{-2}{*}{\cellcolor[HTML]{C0C0C0}\textbf{Sens.}} & \multirow{-2}{*}{\cellcolor[HTML]{C0C0C0}\textbf{Inv.}} & \multirow{-2}{*}{\cellcolor[HTML]{C0C0C0}\textbf{LPD}} & \textbf{Acc.} & \textbf{Pre.} & \textbf{Rec.} & \textbf{pa} & \textbf{hr} \\ \hline
\xmark & \xmark & \xmark & 72.07 & 60.50 & 83.65 & 86.4 & 84.6 \\
\cmark & \xmark & \xmark & 77.02 & 69.82 & 84.21 & 87.4 & 85.1 \\
\xmark & \cmark & \xmark & 76.04 & 67.18 & 84.89 & 87.0 & 85.4 \\
\rowcolor[HTML]{DAE8FC}
\cmark & \cmark & \xmark & 79.74 & 73.74 & 85.72 & 88.8 & 86.2 \\
\xmark & \xmark & \cmark & 73.88 & 63.18 & 84.58 & 86.9 & 86.3 \\
\rowcolor[HTML]{DAE8FC}
\cmark & \cmark & \cmark & \textbf{80.42} & \textbf{74.50} & \textbf{86.34} & \textbf{89.2} & \textbf{87.2} \\ \hline \hline
\end{tabular}%
}
\end{table}

\subsection{Ablation Studies}

\paragraph{Effect of different components.} \cref{tab:ablation_study} shows the performance of Qwen 2.5 Omni \cite{xu2025qwen25omnitechnicalreport} with MoD-DPO++ on AVHBench \cite{sung-bin2025avhbench} and CMM \cite{leng2025the_cmmbenchmark} when different components of the proposed approach are ablated. We report the average metrics across different tasks within each benchmark. Sensitivity, Invariance and LPD are ablated by setting the corresponding hyperparameters $\beta_{\text{sens}}$, $\beta_{\text{inv}}$ and $\gamma_{\text{LPD}}$ to zero. Using LPD increases prediction recall and hallucination resistance significantly, showing that LPD significantly reduces language prior-induced hallucinations.

\begin{table}[]
\centering
\caption{Performance comparison using different forms of rejected samples for MoD-DPO++ on Qwen 2.5 Omni \cite{xu2025qwen25omnitechnicalreport}. $y_l$: rejected text response, $a'$/$v'$: corrupt audio/visual inputs, O: all zeros, RN: gaussian random noise, RAV: random audio/visual input, Diff.: diffusion-based corruption, $t$: steps for diffusion corruption.}
\vspace{-1em}
\label{tab:dpo_negative_sampling}
\resizebox{\columnwidth}{!}{%
\begin{tabular}{ccc|ccc|cc}
\hline\hline
\rowcolor[HTML]{C0C0C0} 
\cellcolor[HTML]{C0C0C0} & \cellcolor[HTML]{C0C0C0} & \cellcolor[HTML]{C0C0C0} & \multicolumn{3}{c|}{\cellcolor[HTML]{C0C0C0}\textbf{AVHBench} \cite{sung-bin2025avhbench}} & \multicolumn{2}{c}{\cellcolor[HTML]{C0C0C0}\textbf{CMM} \cite{leng2025the_cmmbenchmark}} \\ \cline{4-8} 
\rowcolor[HTML]{C0C0C0} 
\multirow{-2}{*}{\cellcolor[HTML]{C0C0C0}\textbf{\begin{tabular}[c]{@{}c@{}}$y_l$\end{tabular}}} & \multirow{-2}{*}{\cellcolor[HTML]{C0C0C0}\textbf{\begin{tabular}[c]{@{}c@{}}$a'$/$v'$\end{tabular}}} & \multirow{-2}{*}{\cellcolor[HTML]{C0C0C0}\textbf{\begin{tabular}[c]{@{}c@{}}$t$\end{tabular}}} & \textbf{Acc.} & \textbf{Pre.} & \textbf{Rec.} & \textbf{pa} & \textbf{hr} \\ \hline
Rel. Opp. & O & - & 77.03 & 69.14 & 84.91 & 87.0 & 86.1 \\
Rel. Opp. & RN & - & 79.93 & 73.98 & 85.87 & 88.8 & \textbf{87.3} \\
Rel. Opp. & RAV & - & 79.03 & 71.87 & 86.18 & 88.4 & 86.3 \\
Rel. Opp. & Diff. & 10 & 72.72 & 62.16 & 83.28 & 86.3 & 84.9 \\
Rel. Opp. & Diff. & 50 & 76.92 & 69.52 & 84.32 & 87.1 & 85.4 \\
\rowcolor[HTML]{DAE8FC}
Rel. Opp. & Diff. & 500 & \textbf{80.42} & \textbf{74.50} & \textbf{86.34} & \textbf{89.2} & 87.2 \\ \hline
Irrelevant & Diff. & 500 & 77.91 & 71.18 & 84.64 & 88.1 & 85.4 \\ \hline\hline
\end{tabular}%
}
\vspace{-2em}
\end{table}
\vspace{-1em}
\paragraph{Choice of rejected and corrupt samples.} \cref{tab:dpo_negative_sampling} shows the performance variation with different forms of rejected (corrupt) samples for MoD-DPO. First, comparing the last two rows in \cref{tab:dpo_negative_sampling}, we can observe that using a rejected response relevant to the opposite modality as $y_l$ leads to significant performance improvement compared to using a completely irrelevant response. We also report the performance of choosing different ways of corrupting audiovisual inputs in the first six rows of \cref{tab:dpo_negative_sampling}. We can observe that diffusing the inputs with noise (similar to VCD \cite{leng2024mitigating_vcd}) results in the best performance. Using a random video/audio and random noise also yields decent performance for the proposed approach. Additionally, using diffusion with a smaller number of steps results in poor performance, as the corrupted inputs appear similar to the original inputs in this case, thereby spuriously shifting the KL divergence in the MoD-DPO loss.

\begin{table*}[]

\parbox{0.73\linewidth}{%
\centering
\resizebox{\linewidth}{!}{
\begin{tabular}{l|ccc|c|cccc}
\hline\hline
\rowcolor[HTML]{C0C0C0} 
\multicolumn{1}{c|}{\cellcolor[HTML]{C0C0C0}} & \multicolumn{3}{c|}{\cellcolor[HTML]{C0C0C0}\textbf{DailyOmni} \cite{zhou2025dailyomni}} & \textbf{MVBench} \cite{li2024mvbench} & \multicolumn{4}{c}{\cellcolor[HTML]{C0C0C0}\textbf{MMAU} \cite{sakshi2025mmau}} \\ \cline{2-9} 
\rowcolor[HTML]{C0C0C0} 
\multicolumn{1}{c|}{\multirow{-2}{*}{\cellcolor[HTML]{C0C0C0}\textbf{Model}}} & \textbf{30s} & \textbf{60s} & \textbf{Avg.} & \textbf{Avg.} & \textbf{Sound} & \textbf{Music} & \textbf{Speech} & \textbf{Avg.} \\ \hline
Qwen 2.5 Omni \cite{xu2025qwen25omnitechnicalreport} & 46.74 & 48.46 & 47.34 & 69.61 & 66.19 & 68.34 & \textbf{59.32} & 64.62 \\
+ DPO \cite{rafailov2023direct_dpo} & 50.95 & 51.93 & 51.44 & {68.21} & 67.23 & 69.99 & 58.34 & 65.19 \\
+ OmniDPO \cite{chen2025omnidpo} & 49.24 & 50.89 & 50.07 & {68.89} & {67.58} & 69.87 & 59.11 & {65.52} \\
\rowcolor[HTML]{DAE8FC}
+ MoD-DPO & 52.45 & 53.54 & 53.00 & {70.95} & \textbf{68.21} & 70.12 & 58.98 & {65.77} \\ 
\rowcolor[HTML]{DAE8FC}
+ MoD-DPO++ & \textbf{53.45} & \textbf{54.18} & \textbf{53.82} & \textbf{71.02} & 68.16 & \textbf{71.60} & {59.24} & \textbf{66.33} \\ \hline
MiniCPM-O 2.6 \cite{yao2024minicpm} & 38.02 & 23.09 & 30.55 & 62.56 & 69.23 & 66.76 & 64.26 & 66.75 \\
+ DPO \cite{rafailov2023direct_dpo} & 39.97 & 28.28 & 34.13 & 63.18 & 69.38 & 67.13 & 64.23 & 66.91 \\
+ OmniDPO \cite{chen2025omnidpo} & 40.68 & 27.98 & 34.33 & \textbf{64.42} & {70.65} & {69.15} & 63.38 & {67.73} \\
\rowcolor[HTML]{DAE8FC}
+ MoD-DPO & 42.86 & 28.56 & 35.71 & 64.15 & \textbf{72.08} & \textbf{70.08} & 64.78 & \textbf{68.98} \\
\rowcolor[HTML]{DAE8FC}
+ MoD-DPO++ & \textbf{43.33} & \textbf{29.87} & \textbf{36.60} & {64.32} & 71.34 & 69.23 & \textbf{64.91} & 68.30 \\\hline\hline
\end{tabular}%
}
\vspace{-1em}
\caption{Comparison of the proposed approach with other DPO methods on general benchmarks: DailyOmni (audiovisual), MVBench (video), and MMAU (audio).}
\label{tab:perf_general}
}
\hfill
\parbox{0.265\linewidth}{%
\centering
\resizebox{\linewidth}{!}{
\includegraphics[width=\linewidth]{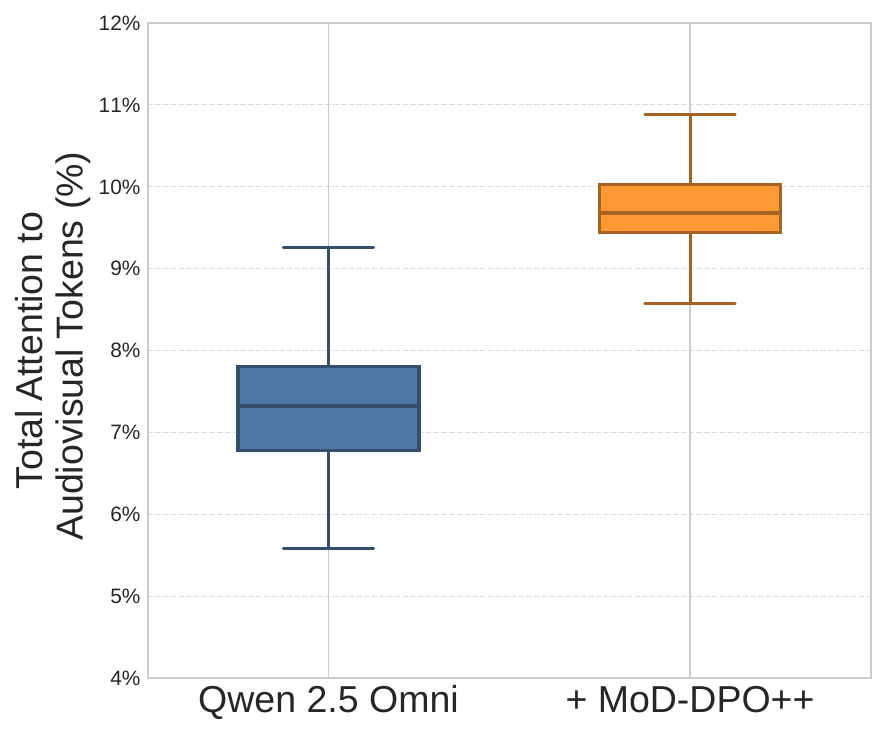}
}
\vspace{-2em}
\captionof{figure}{Total attention to audiovisual tokens as a percentage of total attention to input tokens (including text) on CMM \cite{leng2025the_cmmbenchmark}. }
\label{fig:tpd_attention_plot}
}

\end{table*}

\begin{figure*}[t]
    \centering
    \begin{subfigure}{0.24\linewidth}
        \centering
        \includegraphics[width=\linewidth]{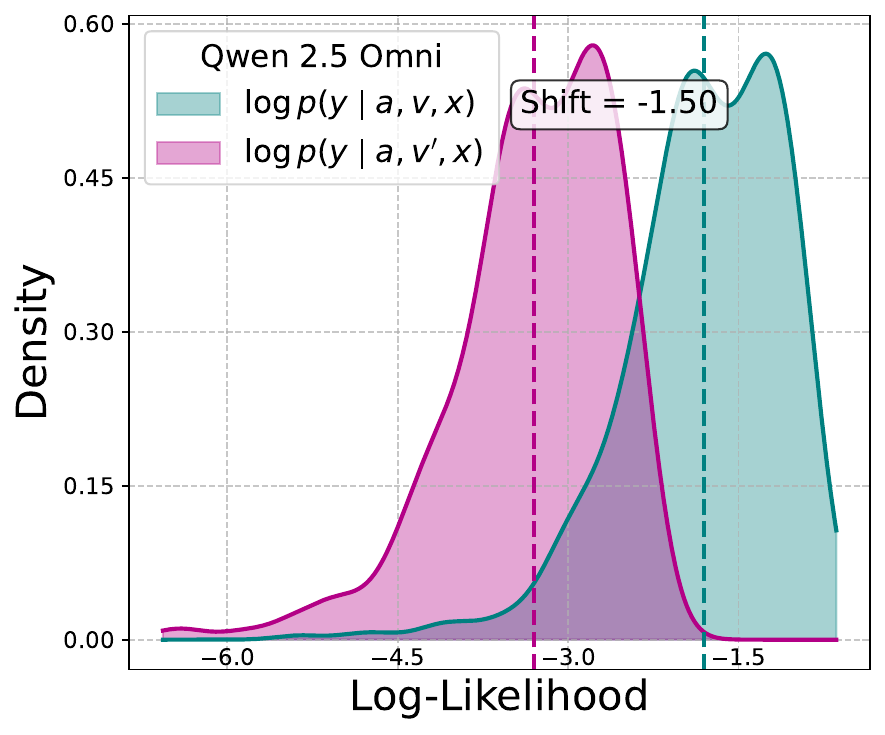}
    \end{subfigure}
    \hfill
    \begin{subfigure}{0.24\linewidth}
        \centering
        \includegraphics[width=\linewidth]{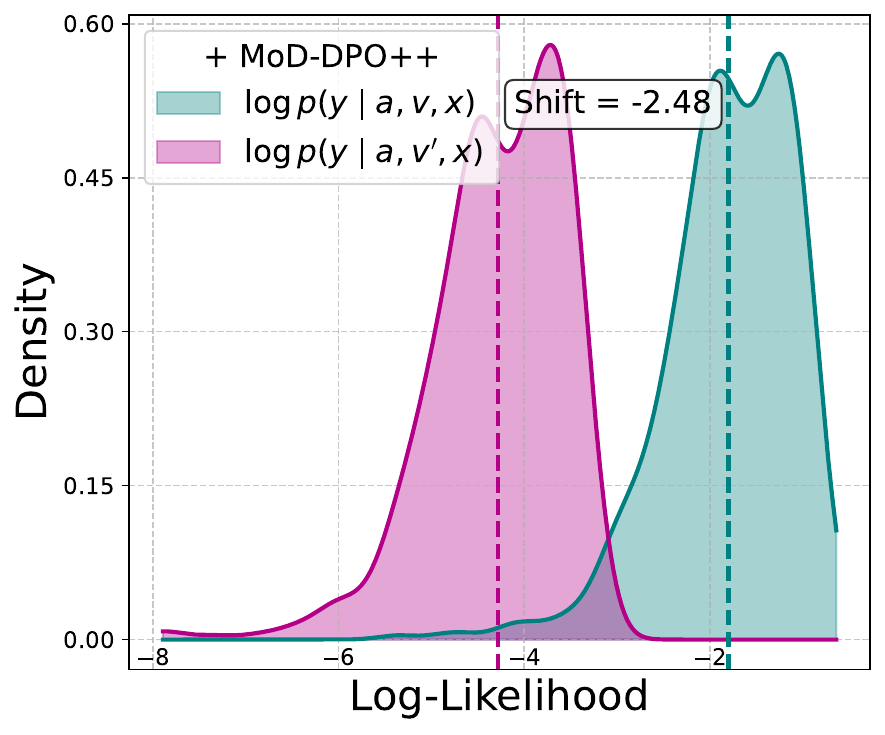}
    \end{subfigure}
    \hfill
    \begin{subfigure}{0.24\linewidth}
        \centering
        \includegraphics[width=\linewidth]{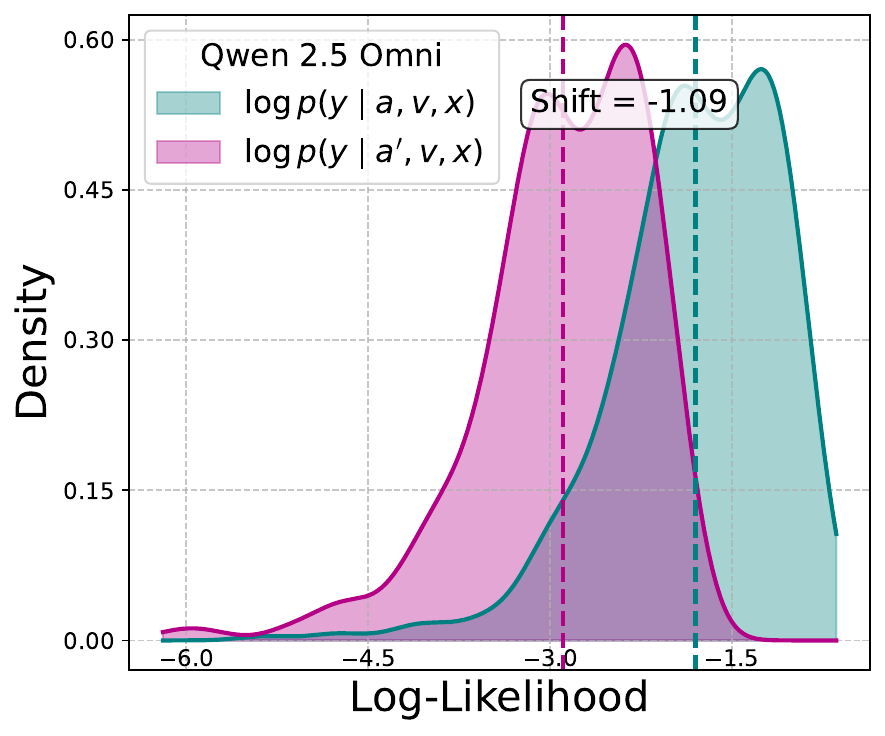}
    \end{subfigure}
    \hfill
    \begin{subfigure}{0.24\linewidth}
        \centering
        \includegraphics[width=\linewidth]{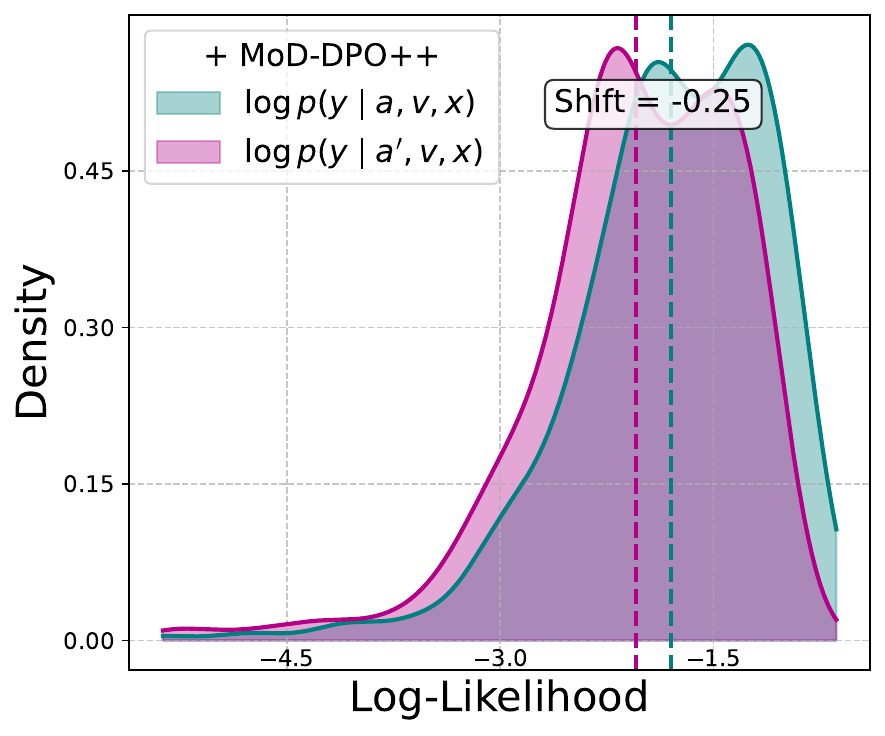}
    \end{subfigure}
    \vspace{-1em}
    \caption{Distribution shift for the log-likelihood of the correct answer when the relevant modality is corrupted (\emph{left two plots}) and when the irrelevant modality is corrupted (\emph{right two plots}) for the audio-driven visual hallucination task in AVHBench \cite{sung-bin2025avhbench}. MoD-DPO++ results in a larger shift when the relevant modality is corrupted (\emph{sensitivity}) and lower shift when the irrelevant modality is corrupted (\emph{invariance}).}
    \vspace{-1em}
    \label{fig:nll_plots_advh}
\end{figure*}

\noindent\textbf{Strength of hyperparameters.} 
\label{subsubsec:strength_hyperparam}
\appref{app_subsec:hyperparam_tuning} shows the effect of varying hyperparameters involved in MoD-DPO++. Increasing $\beta_{\text{sens}}$ up to 0.1 leads to an increase in precision and perception accuracy in AVHBench and CMM, respectively; however, recall and hallucination resistance start dropping sharply after 0.1. For higher strengths of $\beta_{\text{inv}}$ ($>0.02$), model performance deteriorates significantly, which is consistent with the argument in the practical considerations in \cref{subsec:decoupling_modality_inputs}. Additionally, higher strengths of $\gamma_{\text{LPD}}$ result in a sharp decline of precision and perception accuracy, as a heavy penalty on language priors even suppresses the desirable language priors in the backbone LLM.
\begin{table}[]
\centering
\caption{Effect of training with matched/mismatched audiovisual contexts and tasks in the preference training data. Reference is Qwen 2.5 Omni \cite{xu2025qwen25omnitechnicalreport}. A: Audio-related, V: Video-related, AV: Joint audiovisual tasks.}
\vspace{-1em}
\label{tab:match_mismatch_ablation}
\resizebox{\columnwidth}{!}{%
\begin{tabular}{cc|ccc|cc}
\hline\hline
\rowcolor[HTML]{C0C0C0} 
\cellcolor[HTML]{C0C0C0} & \cellcolor[HTML]{C0C0C0} & \multicolumn{3}{c|}{\cellcolor[HTML]{C0C0C0}\textbf{AVHBench}\cite{sung-bin2025avhbench}} & \multicolumn{2}{c}{\cellcolor[HTML]{C0C0C0}\textbf{CMM}\cite{leng2025the_cmmbenchmark}} \\ \cline{3-7} 
\rowcolor[HTML]{C0C0C0} 
\multirow{-2}{*}{\cellcolor[HTML]{C0C0C0}\textbf{AV Context}} & \multirow{-2}{*}{\cellcolor[HTML]{C0C0C0}\textbf{Tasks}} & \textbf{Acc.} & \textbf{Pre.} & \textbf{Rec.} & \textbf{pa} & \textbf{hr} \\ \hline
Mat.  & A,V & 76.69 & 69.29 & 84.09 & 87.6 & 85.8 \\
Mis. & A,V &  78.24 & 72.18 & 84.29 & 88.2 & 86.1 \\
\rowcolor[HTML]{DAE8FC}
Mat.+Mis. & A,V &  \textbf{80.42} & {74.50} & \textbf{86.34} & \textbf{89.2} & {87.2} \\ 
Mat.+Mis. & A,V,\textbf{AV} & 80.33 & \textbf{74.68} & 85.98 & 89.0 & \textbf{87.5} \\ 
\hline\hline
\end{tabular}%
}
\vspace{-1em}
\end{table}

\noindent \textbf{Importance of using mismatched contexts.}
\label{para:importance_of_mismatched_context}
\cref{tab:match_mismatch_ablation} illustrates the importance of using a combination of matched and mismatched audiovisual contexts in the training data proposed in \cref{subsec:dpo_preference_dataset}. Matched context samples contain audio and video from the same source file, while mismatched ones use audio and video from different files. We can observe that using a mix of matched and mismatched audiovisual context results in superior performance.

\noindent \textbf{Adding joint audiovisual data for preference training.} 
\label{para:joint_audiovisual_training}
We include an additional loss function for joint audiovisual tasks as described in \appref{app:audiovisual_dpo_mod_objective} to train on joint audiovisual tasks. We include tasks of audiovisual captioning and audiovisual matching in the preference dataset for this training. \cref{tab:match_mismatch_ablation} (last row) shows the results of including audiovisual tasks, which do not result in significant improvement over the objective with just audio and visual tasks.

\subsection{Robustness to Adversarial Inputs}
Our MoD-DPO++ objective ensures that the output distribution remains robust to changes in the prompt-irrelevant modality (\emph{invariance}), while remaining sensitive to corruptions in the relevant modality. \cref{fig:nll_plots_advh} shows the log-likelihood shifts (using Kernel Density Estimation) for the audio-driven video hallucination task in AVHBench \cite{sung-bin2025avhbench} when the relevant (i.e., video) and irrelevant (i.e., audio) modalities of the input are corrupted. We can observe that the MoD-DPO++ is more invariant (robust) to corruptions in the irrelevant modality and is more sensitive to the corruptions in the relevant modality. 

\subsection{Attention Redistribution to Audiovisual Tokens} To show the efficacy of LPD (\cref{subsec:language_prior_debiasing}), we plot the distribution of combined audiovisual attention as a percent of the total attention to input tokens in generating responses for CMM \cite{leng2025the_cmmbenchmark} for both Qwen 2.5 Omni \cite{xu2025qwen25omnitechnicalreport} and MoD-DPO++ trained models in \cref{fig:tpd_attention_plot}. We take the average over all attention heads and all layers. We can see that with MoD-DPO++, the total attention towards audiovisual tokens increases significantly, indicating that MoD-DPO++ forces the model to focus more on audiovisual inputs.

\subsection{Performance on General Benchmarks} 
\label{subsec:results_general_benchmarks}
\cref{tab:perf_general} shows the improvement in performance of Qwen 2.5 Omni \cite{xu2025qwen25omnitechnicalreport} and MiniCPM-O 2.6 \cite{yao2024minicpm} on general benchmarks. We can see that while the baselines provide inconsistent performance gains, MoD-DPO++ provides consistent gains over all the benchmarks. Additionally, comparison with decode-time approaches is present in \appref{app_subsec:decode_time_comparison} and multi-turn results are present in \appref{app_subsec:multi_turn_omni}.


\section{Discussion \& Conclusion}

We presented MoD-DPO, a modality-decoupled preference optimization framework that mitigates cross-modal hallucinations in omni LLMs by enforcing invariance to irrelevant modality corruption and sensitivity to perturbations in the task-relevant modality. Combined with a language-prior debiasing penalty and a large automatically generated preference dataset, MoD-DPO consistently improves robustness across challenging audiovisual benchmarks. While the current framework does not explicitly promote beneficial cross-modal synergy for tasks requiring complementary audiovisual reasoning (discussed further in \appref{app_sec:cross_modal_discussion}), our findings underscore the value of structured preference optimization for reliable omni-modal systems.


\section*{Acknowledgements}

Research was sponsored by the Army Research Office and was accomplished under Cooperative Agreement Number W911NF-25-2-0040. The views and conclusions contained in this document are those of the authors and should not be interpreted as representing the official policies, either expressed or implied, of the Army Research Office or the U.S. Government. The U.S. Government is authorized to reproduce and distribute reprints for Government purposes notwithstanding any copyright notation herein.
{
    \small
    \bibliographystyle{ieeenat_fullname}
    \bibliography{main}
}

\clearpage
\onecolumn
\setcounter{page}{1}

\begin{center}
{\Large \bf MoD-DPO: Towards Mitigating Cross-modal Hallucinations in Omni LLMs using Modality Decoupled Preference Optimization} \\[0.5cm]
{\large  Supplementary Material}
\end{center}

\vspace{1cm}

\appendix

{\noindent \Large {Table of Contents}
}

\begin{itemize}
    \item Methodological Details \dotfill \appref{app_sec:methodological_details}
    \begin{itemize}
        \item {\small {Finding the Optimal Policy}} \dotfill \appref{app:opt_policy_proof}
        \item {\small {Reward Function for Optimal Policy}} \dotfill \appref{app:reward_proof}
        \item {\small {Decoupled Loss Function}} \dotfill \appref{app:bt_loss_proof}
        \item {\small {Symmetric Objective for Audio-Related Prompts}} \dotfill \appref{app:audio_objective}
        \item {\small {(Optional) Objective for Prompts requiring both audio and visual information}} \dotfill \appref{app:audiovisual_dpo_mod_objective}
        \item Training compute details \dotfill \appref{app_subsec:training_compute_details}
    \end{itemize}
    \item Preference Data Details \dotfill \appref{app:preference_data}
    \begin{itemize}
        \item {\small {Detailed Pipeline}} \dotfill \appref{app_subsec:detailed_data_stages}
        \item {\small {Preference Data Samples}} \dotfill \appref{app_subsec:pref_data_samples}
    \end{itemize}
    \item Experimental Details \dotfill \appref{app_sec:experimental_details}
    \begin{itemize}
        \item {\small {Evaluation Metrics}} \dotfill \appref{app_subsec:evaluation_metrics}
        \item {\small {Evaluation Protocol}} \dotfill \appref{app_subsec:eval_protocol}
        \item {\small {Baseline Implementations}} \dotfill \appref{app_subsec:baseline_implementations}
    \end{itemize}
    \item Additional Results \dotfill \appref{app_sec:additional_results}
    \begin{itemize}
        \item {\small {Hyper-parameter Tuning}} \dotfill \appref{app_subsec:hyperparam_tuning}
        \item {\small {Comparison with decode-time approaches}} \dotfill \appref{app_subsec:decode_time_comparison}
        \item {\small {Multi-turn omni-modal results}} \dotfill \appref{app_subsec:multi_turn_omni}
    \end{itemize}
    \item Discussion on Cross-Modal Synergy \dotfill \appref{app_sec:cross_modal_discussion}
    \item Prompt Pool \dotfill \appref{app:prompt_pool}
\end{itemize}

\section{Methodological Details}
\label{app_sec:methodological_details}

\subsection{Finding the Optimal Policy}
\label{app:opt_policy_proof}

Consider the objective in \cref{eq:dpo_mod} for prompts that are related to the visual modality. For clarity, define $p_\theta \triangleq \pi_\theta(y \mid a,v,x^v)$, $p_{\text{ref}} \triangleq \pi_{\text{ref}}(y \mid a,v,x^v)$, $q_{\text{inv}}(y) \triangleq \pi'_\theta(y \mid a',v,x^v)$, and $q_{\text{sens}}(y) \triangleq \pi'_\theta(y \mid a,v',x^v)$. 
We also assume that $q_{\mathrm{inv}}$ and $q_{\mathrm{sens}}$ are treated as fixed target distributions within a single optimization step (that is, they do not depend on the parameters with respect to which we differentiate) as described in \cref{subsec:decoupling_modality_inputs}.

With these assumptions, and suppressing the conditioning variables $(a,v,x^v)$ for brevity, the per-\,$(a,v,x^v)$ objective can be written as,
\begin{equation}
\small
\label{eq:app_obj}
\mathcal{J}(p_\theta)
= \sum_{y} p_\theta(y)\, r(y)
- \beta \mathbb{D}_{\text{KL}}\!\big(p_\theta \,\|\, p_{\mathrm{ref}}\big)
- \beta_{\mathrm{inv}} \mathbb{D}_{\text{KL}}\!\big(p_\theta \,\|\, q_{\mathrm{inv}}\big)
+ \beta_{\mathrm{sens}} \mathbb{D}_{\text{KL}}\!\big(p_\theta \,\|\, q_{\mathrm{sens}}\big),
\end{equation}
subject to $p_\theta(y) \ge 0$ and $\sum_{y} p_\theta(y)=1$.
We can find the optimal policy that maximizes \cref{eq:app_obj} using Lagrange's method. We introduce a Lagrange multiplier $\lambda$ for the normalization constraint, and expand the KL divergence terms to obtain,
\begin{equation}
\begin{aligned}
\mathcal{J}'(p_\theta,\lambda)
&= \sum_{y} p_\theta(y) r(y)
- \beta \sum_{y} p_\theta(y) \log \frac{p_\theta(y)}{p_{\mathrm{ref}}(y)}
- \beta_{\mathrm{inv}} \sum_{y} p_\theta(y) \log \frac{p_\theta(y)}{q_{\mathrm{inv}}(y)} \\
&\quad
+ \beta_{\mathrm{sens}} \sum_{y} p_\theta(y) \log \frac{p_\theta(y)}{q_{\mathrm{sens}}(y)}
+ \lambda \!\left(\sum_{y} p_\theta(y) - 1\right).
\end{aligned}
\label{eq:app_lagrangian}
\end{equation}

\noindent To obtain the stationary condition, we take the partial derivative of $\mathcal{J}'$ with respect to $p_\theta(y)$ and set it to zero,
\begin{equation}
\small
\begin{aligned}
0
&= \frac{\partial \mathcal{L}}{\partial p_\theta(y)}
= r(y)
- \beta \Big(\log p_\theta(y) - \log p_{\mathrm{ref}}(y) + 1\Big)
- \beta_{\mathrm{inv}} \Big(\log p_\theta(y) - \log q_{\mathrm{inv}}(y) + 1\Big) \\
&\quad
+ \beta_{\mathrm{sens}} \Big(\log p_\theta(y) - \log q_{\mathrm{sens}}(y) + 1\Big)
+ \lambda.
\end{aligned}
\label{eq:app_stationarity_raw}
\end{equation}
We can now group the terms by $\log p_\theta(y)$ and collect the constants. Additionally, we define $\tau \triangleq \beta + \beta_{\mathrm{inv}} - \beta_{\mathrm{sens}}$ for simplicity. Then \cref{eq:app_stationarity_raw} becomes,
\begin{align}
0
&= r(y)
- \tau \log p_\theta(y)
+ \beta \log p_{\mathrm{ref}}(y)
+ \beta_{\mathrm{inv}} \log q_{\mathrm{inv}}(y)
- \beta_{\mathrm{sens}} \log q_{\mathrm{sens}}(y)
+ \big(\lambda - \beta - \beta_{\mathrm{inv}} + \beta_{\mathrm{sens}}\big).
\label{eq:app_stationarity_grouped}
\end{align}
Finally, we can arrange the above equation to isolate $\log p_\theta(y)$ as,
\begin{equation}
\label{eq:app_logp}
\log p_\theta(y)
= \frac{1}{\tau}\Big(
r(y)
+ \beta \log p_{\mathrm{ref}}(y)
+ \beta_{\mathrm{inv}} \log q_{\mathrm{inv}}(y)
- \beta_{\mathrm{sens}} \log q_{\mathrm{sens}}(y)
\Big)
+ C,
\end{equation}
where the constant $C$ absorbs $\lambda - \beta - \beta_{\mathrm{inv}} + \beta_{\mathrm{sens}}$ and enforces normalization.

\noindent Exponentiating \cref{eq:app_logp} yields, up to a normalization constant,
\begin{equation}
\label{eq:app_p_propto}
p_\theta(y)
\propto
\exp\!\big(r(y)/\tau\big)\,
p_{\mathrm{ref}}(y)^{\beta/\tau}\,
q_{\mathrm{inv}}(y)^{\beta_{\mathrm{inv}}/\tau}\,
q_{\mathrm{sens}}(y)^{-\beta_{\mathrm{sens}}/\tau}
\end{equation}
\begin{equation}
\implies \pi^\star_\theta(y \mid a, v, x^v)
\propto
\exp\!\big(r(a,v,x^v,y)/\tau\big)\,
\pi_{\mathrm{ref}}(y \mid a,v,x^v)^{\beta/\tau}\,
\pi'_\theta(y \mid a',v,x^v)^{\beta_{\mathrm{inv}}/\tau}\,
\pi'_\theta(y \mid a,v',x^v)^{-\beta_{\mathrm{sens}}/\tau}
\end{equation}
which is exactly \cref{eq:dpo_mod_opt_policy}.


\subsection{Reward Function for Optimal Policy}
\label{app:reward_proof}

Starting from \cref{eq:dpo_mod_opt_policy}, we take the natural logarithm of both sides and isolate $r(a,v,x^v,y)$. Absorbing the normalization constant into a function $W(a,v,x^v)$ that does not depend on $y$ gives,
\begin{align}
\tau \log \pi_\theta(y \mid a,v,x^v)
&= r(a,v,x^v,y)
+ \beta \log \pi_{\mathrm{ref}}(y \mid a,v,x^v) \nonumber\\
&\quad
+ \beta_{\mathrm{inv}} \log \pi'_\theta(y \mid a',v,x^v)
- \beta_{\mathrm{sens}} \log \pi'_\theta(y \mid a,v',x^v)
+ W(a,v,x^v),
\end{align}
which, upon rearranging yields,
\begin{align}
r(a,v,x^v,y)
&= \tau \log \pi_\theta(y \mid a,v,x^v)
- \beta \log \pi_{\mathrm{ref}}(y \mid a,v,x^v) \nonumber\\
&\quad
- \beta_{\mathrm{inv}} \log \pi'_\theta(y \mid a',v,x^v)
+ \beta_{\mathrm{sens}} \log \pi'_\theta(y \mid a,v',x^v)
+ W(a,v,x^v),
\end{align}
which is \cref{eq:reward_dpo_mod}. The function $W(a,v,x^v)$ is determined by enforcing that $\sum_{y} \pi^\star_\theta(y \mid a,v,x^v)=1$ and does not affect optimization under pairwise preference models, since it cancels in likelihood ratios.


\subsection{Decoupled Loss Function}
\label{app:bt_loss_proof}

For a given a triple $((a,v,x^v), y_w, y_l)$ and the reward $r(\cdot)$ in \cref{eq:reward_dpo_mod}, the Bradley Terry model assigns the probability that $y_w$ is preferred over $y_l$ as
\[
\Pr\!\big[y_w \succ y_l \,\big|\, a,v,x^v\big]
= \sigma\!\Big(r(a,v,x^v,y_w) - r(a,v,x^v,y_l)\Big),
\]
where $\sigma(t)=1/(1+e^{-t})$ is the sigmoid function.
Using \cref{eq:reward_dpo_mod}, the difference
$r(a,v,x^v,y_w) - r(a,v,x^v,y_l)$
cancels the function $W(a,v,x^v)$ and yields
\begin{align}
&\tau \Big(\log \pi_\theta(y_w \mid a,v,x^v) - \log \pi_\theta(y_l \mid a,v,x^v)\Big)
- \beta \Big(\log \pi_{\mathrm{ref}}(y_w \mid a,v,x^v) - \log \pi_{\mathrm{ref}}(y_l \mid a,v,x^v)\Big) \nonumber\\
&\quad
- \beta_{\mathrm{inv}} \Big(\log \pi'_\theta(y_w \mid a',v,x^v) - \log \pi'_\theta(y_l \mid a',v,x^v)\Big)
+ \beta_{\mathrm{sens}} \Big(\log \pi'_\theta(y_w \mid a,v',x^v) - \log \pi'_\theta(y_l \mid a,v',x^v)\Big).
\label{eq:app_reward_diff}
\end{align}
Taking the negative expected log-likelihood over the preference dataset $\mathcal{D}^{\mathrm{pref}}_{\mathrm{text}}$ yields \cref{eq:dpo_mod_vision_final}.


\subsection{Symmetric Objective for Audio-Related Prompts}
\label{app:audio_objective}

For prompts $x^a$ that are related to the audio modality, the roles of audio and visual inputs are exchanged. The objective in ~\cref{eq:dpo_mod} becomes,
\begin{align}
\max_{\pi_\theta}\;
&\mathbb{E}\Big[r(a,v,x^a,y)\Big]
- \beta\, \mathbb{D}_{\text{KL}}\!\Big(\pi_\theta(\cdot \mid a,v,x^a)\,\big\|\,\pi_{\mathrm{ref}}(\cdot \mid a,v,x^a)\Big) \nonumber\\
&\quad
- \beta_{\mathrm{inv}}\, \mathbb{D}_{\text{KL}}\!\Big(\pi_\theta(\cdot \mid a,v,x^a)\,\big\|\,\pi_\theta(\cdot \mid a,v',x^a)\Big) \nonumber\\
&\quad
+ \beta_{\mathrm{sens}}\, \mathbb{D}_{\text{KL}}\!\Big(\pi_\theta(\cdot \mid a,v,x^a)\,\big\|\,\pi_\theta(\cdot \mid a',v,x^a)\Big),
\end{align}
where invariance now penalizes deviations when the irrelevant modality (visual) is corrupted and sensitivity rewards deviations when the relevant modality (audio) is corrupted. Repeating the derivations in \appref{app:opt_policy_proof} and \appref{app:reward_proof} with $a$ and $v$ exchanged yields the audio counterpart of \cref{eq:dpo_mod_vision_final}:
\begin{align}
\mathcal{L}_{\mathrm{MoD}}^{a}
&= -\mathbb{E}\Bigg[
\log \sigma\!\Bigg( \tau
\log \frac{\pi_{\theta}(y_w \mid a,v,x^a)}{\pi_{\theta}(y_l \mid a,v,x^a)}
-\beta \log \frac{\pi_{\mathrm{ref}}(y_w \mid a,v,x^a)}{\pi_{\mathrm{ref}}(y_l \mid a,v,x^a)} \nonumber\\[-3pt]
&\qquad
-\beta_{\mathrm{inv}} \log \frac{\pi'_{\theta}(y_w \mid a,v',x^a)}{\pi'_{\theta}(y_l \mid a,v',x^a)}
+\beta_{\mathrm{sens}} \log \frac{\pi'_{\theta}(y_w \mid a',v,x^a)}{\pi'_{\theta}(y_l \mid a',v,x^a)}
\Bigg)
\Bigg].
\end{align}
Consequently, the final loss $\mathcal{L}_{\mathrm{MoD}} = \mathcal{L}_{\mathrm{MoD}}^{v} + \mathcal{L}_{\mathrm{MoD}}^{a}$ in \cref{eq:loss_dpo_mod_decoupled} follows.

\subsection{(Optional) Objective for Prompts requiring both audio and visual information}
\label{app:audiovisual_dpo_mod_objective}

For prompts $x^{av}$ that require both audio and visual information for responding, we drop the \emph{invariance} term in \cref{eq:dpo_mod}. The objective in ~\cref{eq:dpo_mod} becomes,
\begin{align}
\max_{\pi_\theta}\;
&\mathbb{E}\Big[r(a,v,x^{av},y)\Big]
- \beta\, \mathbb{D}_{\text{KL}}\!\Big(\pi_\theta(\cdot \mid a,v,x^{av})\,\big\|\,\pi_{\mathrm{ref}}(\cdot \mid a,v,x^{av})\Big) \nonumber\\
&\quad
+ \beta_{\mathrm{sens}}\, \mathbb{D}_{\text{KL}}\!\Big(\pi_\theta(\cdot \mid a,v,x^{av})\,\big\|\,\pi_\theta(\cdot \mid a',v',x^{av})\Big),
\end{align}
and the corresponding loss function for audiovisual prompts becomes the following,
\begin{align}
\mathcal{L}_{\mathrm{MoD}}^{av}
&= -\mathbb{E}\Bigg[
\log \sigma\!\Bigg( \tau
\log \frac{\pi_{\theta}(y_w \mid a,v,x^{av})}{\pi_{\theta}(y_l \mid a,v,x^{av})}
-\beta \log \frac{\pi_{\mathrm{ref}}(y_w \mid a,v,x^{av})}{\pi_{\mathrm{ref}}(y_l \mid a,v,x^{av})} \nonumber\\[-3pt]
&\qquad
+\beta_{\mathrm{sens}} \log \frac{\pi'_{\theta}(y_w \mid a',v',x^{av})}{\pi'_{\theta}(y_l \mid a',v',x^{av})}
\Bigg)
\Bigg].
\end{align}

\begin{figure*}[t]
\centering

\begin{minipage}[t]{0.55\textwidth}
\vspace{0pt}
\centering

\captionof{table}{Training compute for baselines and MoD-DPO variants.}
\label{tab:compute_comparison}
\vspace{-0.5em}

\resizebox{\textwidth}{!}{%
\begin{tabular}{c|c|ccccc|ccc}
\hline \hline
\rowcolor[HTML]{C0C0C0} 
 &  & \multicolumn{4}{c|}{\textbf{Per Training Iteration}} 
 & \multicolumn{3}{c}{\textbf{Total till Train. Converg.}} \\ \cline{3-9} 

\rowcolor[HTML]{C0C0C0} 
 &  & \multicolumn{2}{c}{\textbf{\# Fwd.}} 
 & \multicolumn{2}{c|}{\textbf{\# Back.}}  
 &  &  &  \\

\rowcolor[HTML]{C0C0C0} 
\multirow{-3}{*}{\textbf{Method}} 
& \multirow{-3}{*}{\textbf{\begin{tabular}[c]{@{}c@{}}Data \\ Preproc.\\ (Offline)\end{tabular}}} 
& $\pi_\theta$ & $\pi_\text{ref}$ 
& $\pi_\theta$ & $\pi_\text{ref}$ 
& \textbf{\# hrs} 
& \textbf{\# iters} 
& \textbf{\begin{tabular}[c]{@{}c@{}}FLOPS$\downarrow$\\ ($\times 10^{18}$)\end{tabular}} \\ 
\hline

Vanila DPO & - & 2 & 2 & 2 & 0  & 4.3 & 5k & 13.75 \\
OmniDPO & a', v' & 4 & 4 & 4 & 0  & 3.0 & 3k & 16.53 \\

\rowcolor[HTML]{DAE8FC}
MoD-DPO & a', v' & 6 & 2 & 2 & 0  & 2.0 & 1.8k & \underline{7.43} \\

\rowcolor[HTML]{DAE8FC}
MoD-DPO++ & a', v' & 6 & 4 & 2 & 0  & 1.8 & 1.5k & \textbf{7.23} \\

\hline \hline
\end{tabular}
}

\end{minipage}
\hfill
\begin{minipage}[t]{0.44\textwidth}
\vspace{0pt}
\centering

\includegraphics[width=\linewidth]{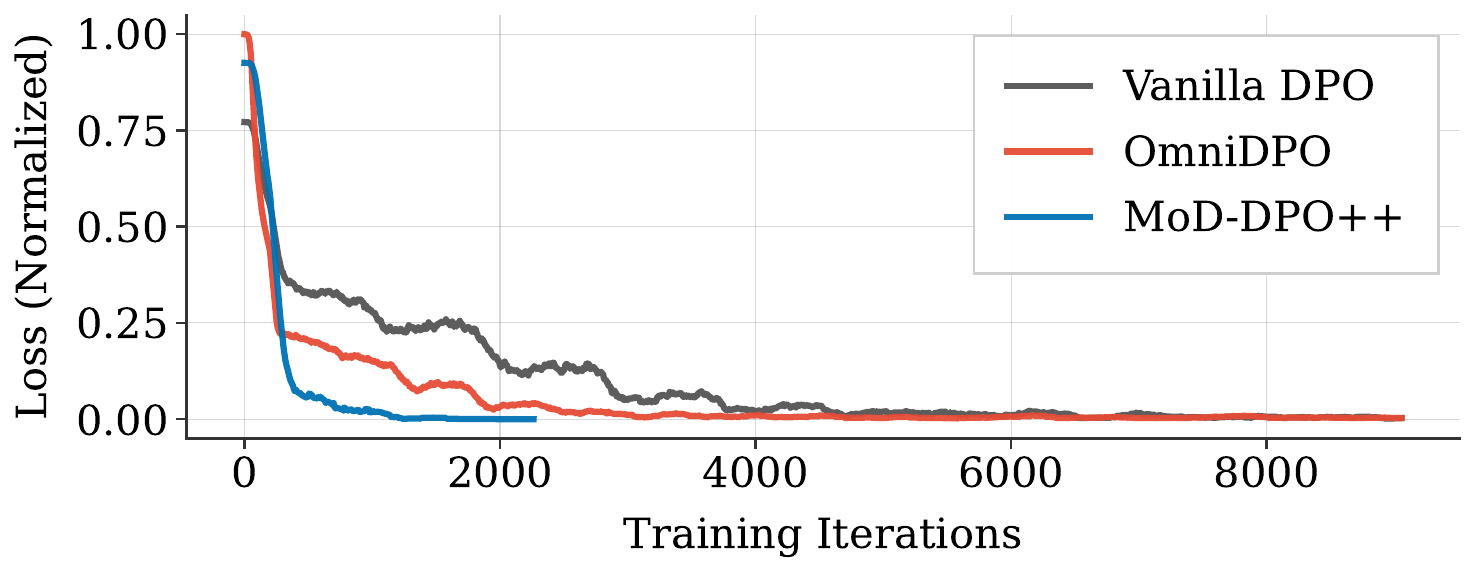}

\vspace{-0.5em}
\caption{Training loss of baselines and MoD-DPO++.}
\label{fig:training_loss}

\end{minipage}

\vspace{-0.5em}
\end{figure*}

\subsection{Training compute details}
\label{app_subsec:training_compute_details}

We report the detailed compute comparison of MoD-DPO with Vanilla DPO \cite{rafailov2023direct_dpo} and OmniDPO \cite{chen2025omnidpo} in \cref{tab:compute_comparison}. As noted in \cref{subsec:language_prior_debiasing}, MoD-DPO++ adds forward passes through the policy model $\pi_\theta$. Compared to OmniDPO, it uses the same data augmentation (hence same data processing compute) but two extra forward passes for policy $\pi_\theta$ and reference $\pi_{\text{ref}}$ (see \cref{tab:compute_comparison}). For MoD-DPO++, since $\pi_\theta'$ in \cref{eq:dpo_mod_opt_policy}  is treated as fixed (see \cref{subsec:decoupling_modality_inputs}), no gradients are required, resulting in fewer backward passes and lower FLOPs per iteration. As shown in \cref{fig:training_loss,tab:compute_comparison}, MoD-DPO++ converges nearly twice as fast as OmniDPO, and all models are trained to convergence.

\section{Preference Data Details}
\label{app:preference_data}

\subsection{Detailed Pipeline}
\label{app_subsec:detailed_data_stages}

\subsubsection{Stage - 1} 

\begin{figure}[!h]
    \centering
    \includegraphics[width=0.4\linewidth]{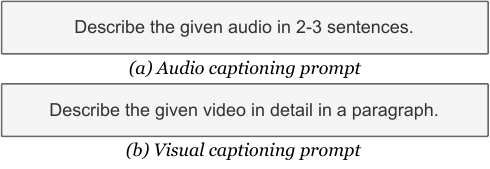}
    \caption{\textbf{Audio and visual captioning prompts} used in Stage-1 of Preference Data Generation \cref{subsec:dpo_preference_dataset}. Audio captions are obtained from AudioFlamingo 3 \cite{ghosh2025audioflamingo3_af3} and visual captions are obtained from GPT-4o \cite{hurst2024gpt4o}.}
    \label{fig:audio_visual_caption_prompt}
\end{figure}

We use the prompts present in \cref{fig:audio_visual_caption_prompt} to obtain audio prompts from AudioFlamingo 3 \cite{ghosh2025audioflamingo3_af3} and the visual prompts from GPT-4o \cite{hurst2024gpt4o}. For audio captioning, we pass the entire audio to the AudioFlamingo 3 model and for visual captioning, we pass frames at a frame rate of 1 frame per second to GPT-4o for visual captioning. Additionally, we use RAM++ \cite{huang2025_rampp} to obtain the visual tags present in the video by passing frames from the video at 1 FPS to the RAM++ model, and we take a union of all the tags obtained as the objects/events present in the video. 

\subsubsection{Stage - 2} 
For the captioning task, Stage-2 involves associating the prompts \emph{``Describe only the audio in detail."} and \emph{``Describe only the video in detail."} to the audio and visual captions obtained in Stage-1, respectively. For the object/event presence subtask, there are three sub-stages involved. 

\begin{figure}
    \centering
    \includegraphics[width=\linewidth]{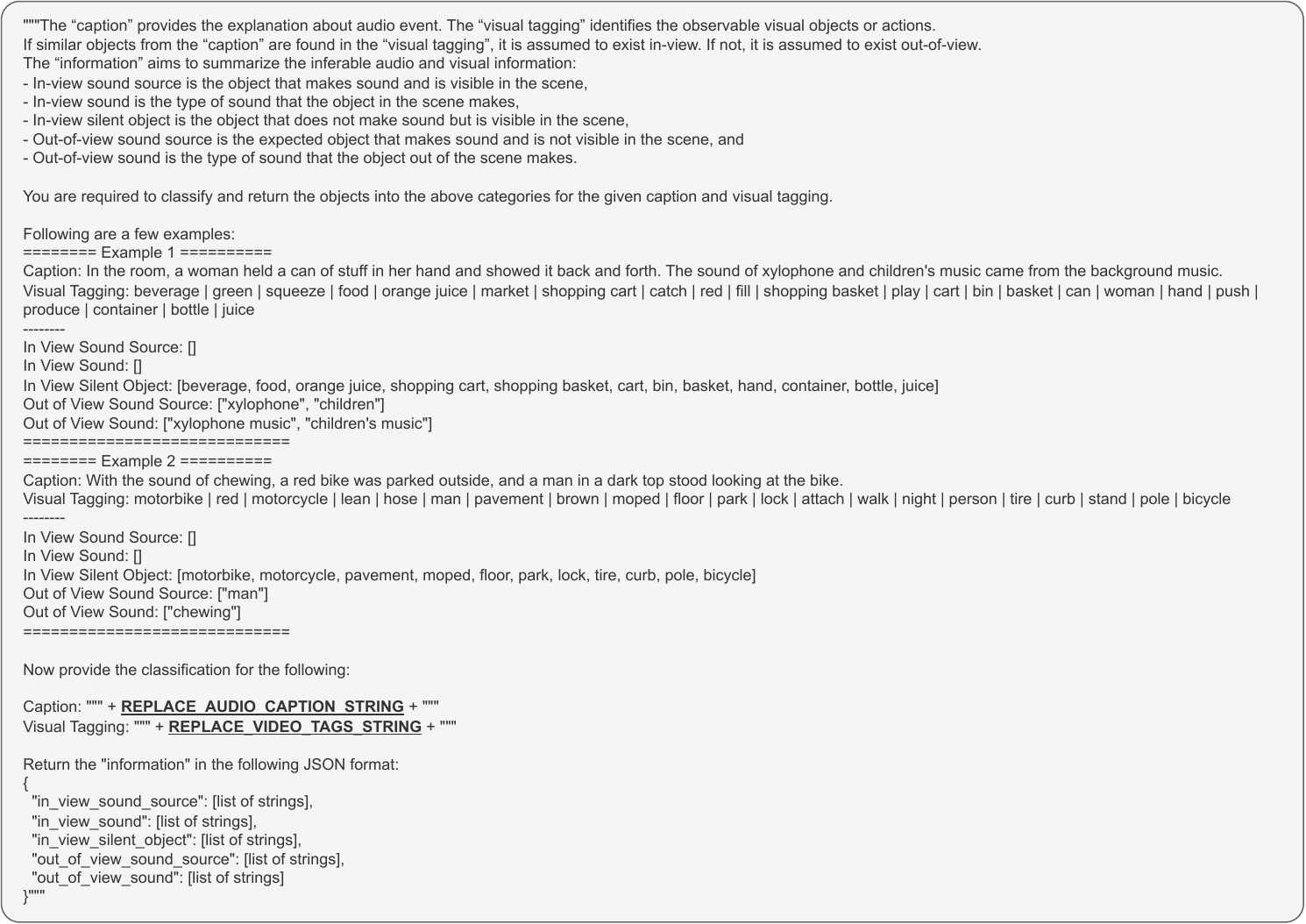}
    \caption{\textbf{QA Generation Prompt - Stage 2.1.} used for classifying the events/objects in the current audiovisual input into different classes similar to AVHBench \cite{sung-bin2025avhbench}.}
    \label{fig:qa_generation_stage_2_1_prompt}
\end{figure}

First, we pass the audio and visual information obtained in the form of audio/visual captions and video tags to GPT-4o using prompts in \cref{fig:qa_generation_stage_2_1_prompt} and classify different objects/events present in the inputs to the following categories -- \emph{``in-view sound source"}, \emph{``in-view sound"}, \emph{``in-view silent object"}, \emph{``out-of-view sound source"}, and \emph{``out-of-view sound"}. This classification is similar to that of AVHBench \cite{sung-bin2025avhbench} and simplifies the automatic generation of QA pairs. 

Next, we automatically generate the QA pairs using the following template -- \emph{``Is the \{object/event\} making sound in the audio?"} for audio presence (\emph{video-driven audio hallucination}) tasks and \emph{``Is the \{object/event\} visible in the video?"} for visual presence (\emph{audio-driven visual hallucination}) tasks. For the visual presence tasks, objects/events classified as \emph{``in-view sound source"} are assigned an answer of \emph{``Yes"}, and \emph{``out-of-view sound source"} \& \emph{``out-of-view sounds"} are assigned an answer of \emph{``No"}. For the audio presence tasks, objects/events classified as \emph{``in-view sound source"} \& \emph{``in-view sound"} are assigned an answer of \emph{``Yes"} and \emph{``in-view silent objects"} are assigned an answer of \emph{``No"}.

\begin{figure}
    \centering
    \includegraphics[width=\linewidth]{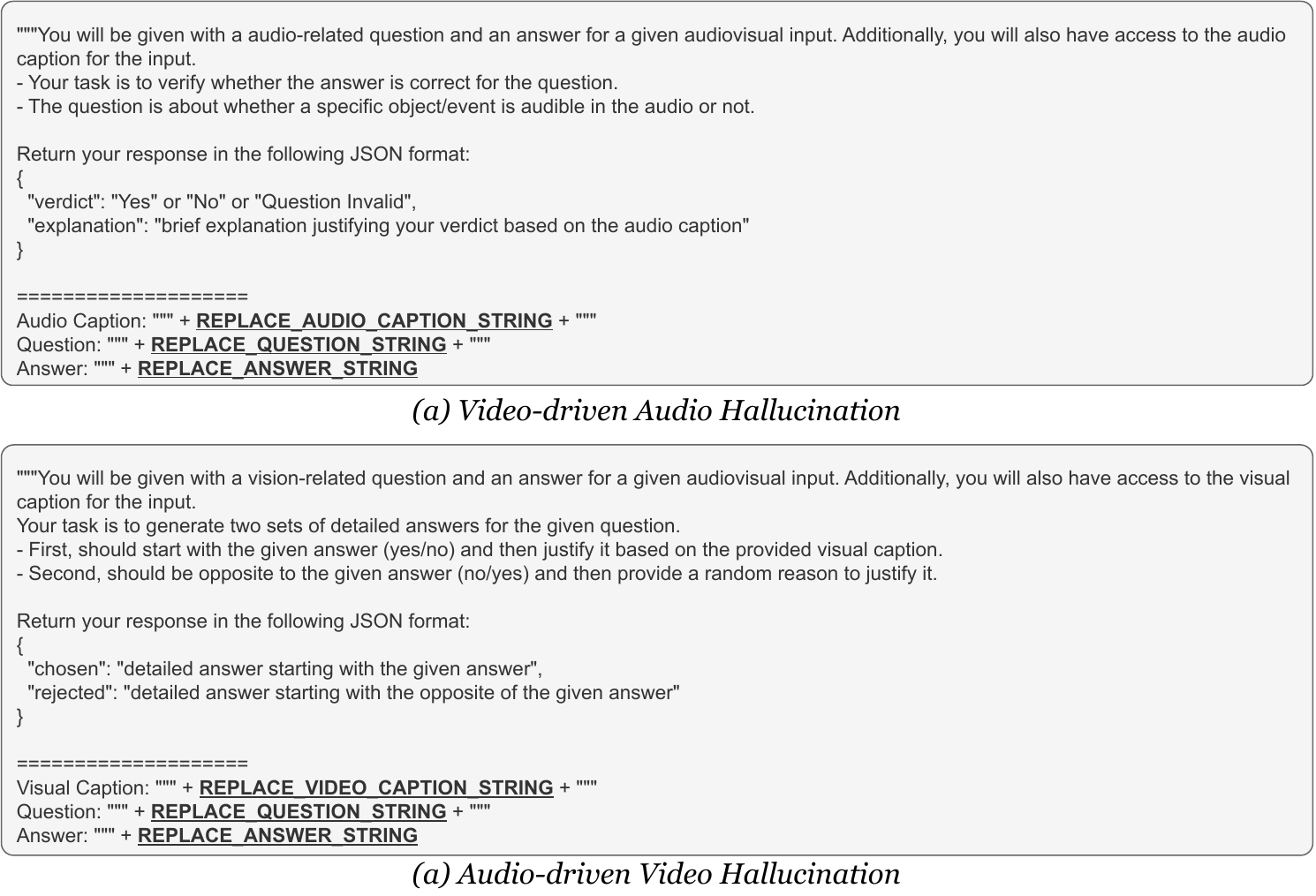}
    \caption{\textbf{QA Generation Verification Prompt - Stage 2.3.} used to verify the automatic QAs generated for video-driven audio hallucination and audio-driven video hallucination.}
    \label{fig:qa_generation_stage_2_3_prompt}
\end{figure}

Finally, since the data generation pipeline in the above paragraph is automatic, we run a round of automatic verification through GPT-4o using prompts in \cref{fig:qa_generation_stage_2_3_prompt} to ensure that the generated data are indeed consistent and correct.

\subsubsection{Stage - 3}

\begin{figure}
    \centering
    \includegraphics[width=\linewidth]{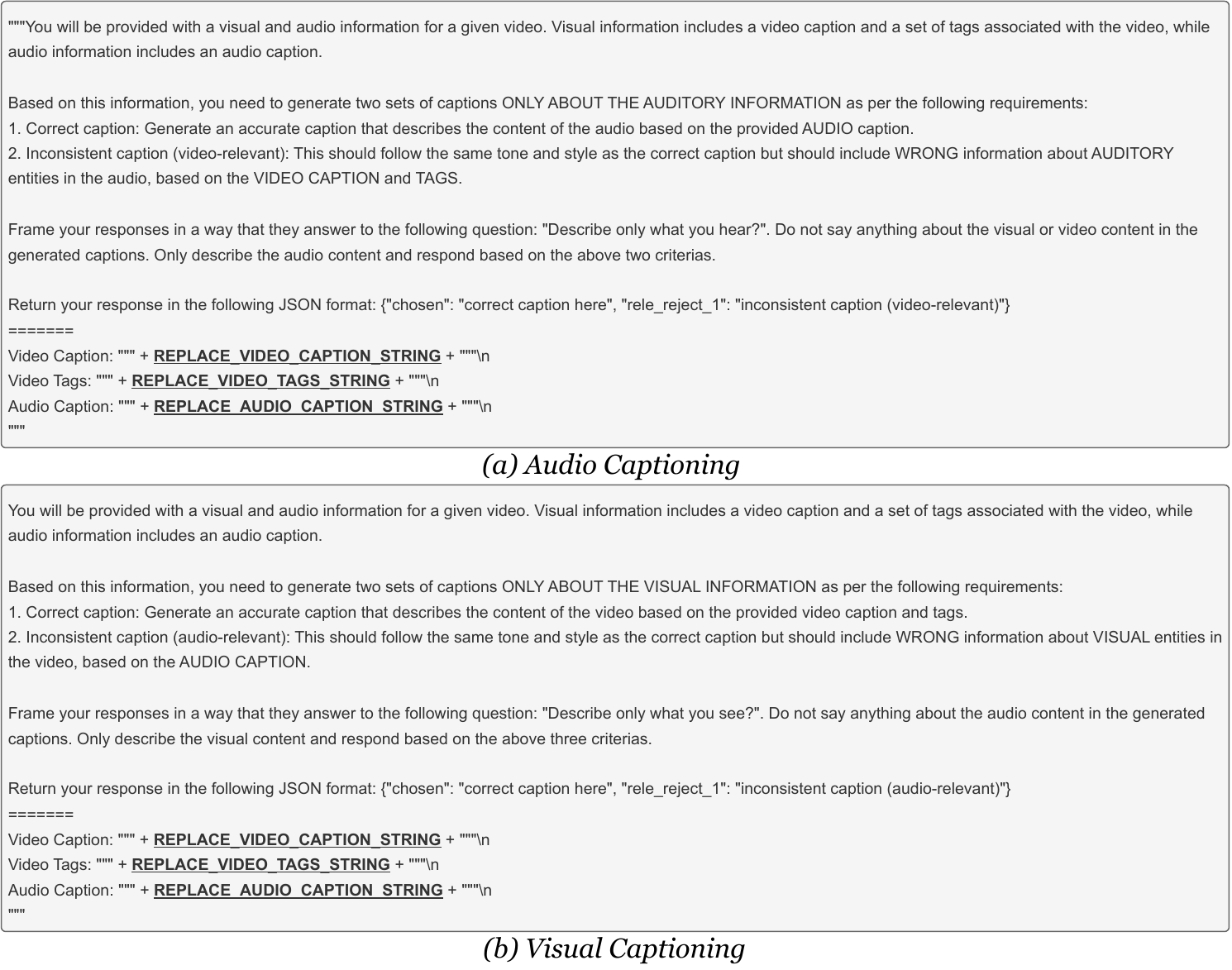}
    \caption{\textbf{Preference Data Generation Prompt - Stage 3 (Captioning)} used to generate preference data pairs for the captioning task for training MoD-DPO.}
    \label{fig:dpo_stage3_caption_prompt}
\end{figure}

\begin{figure}
    \centering
    \includegraphics[width=\linewidth]{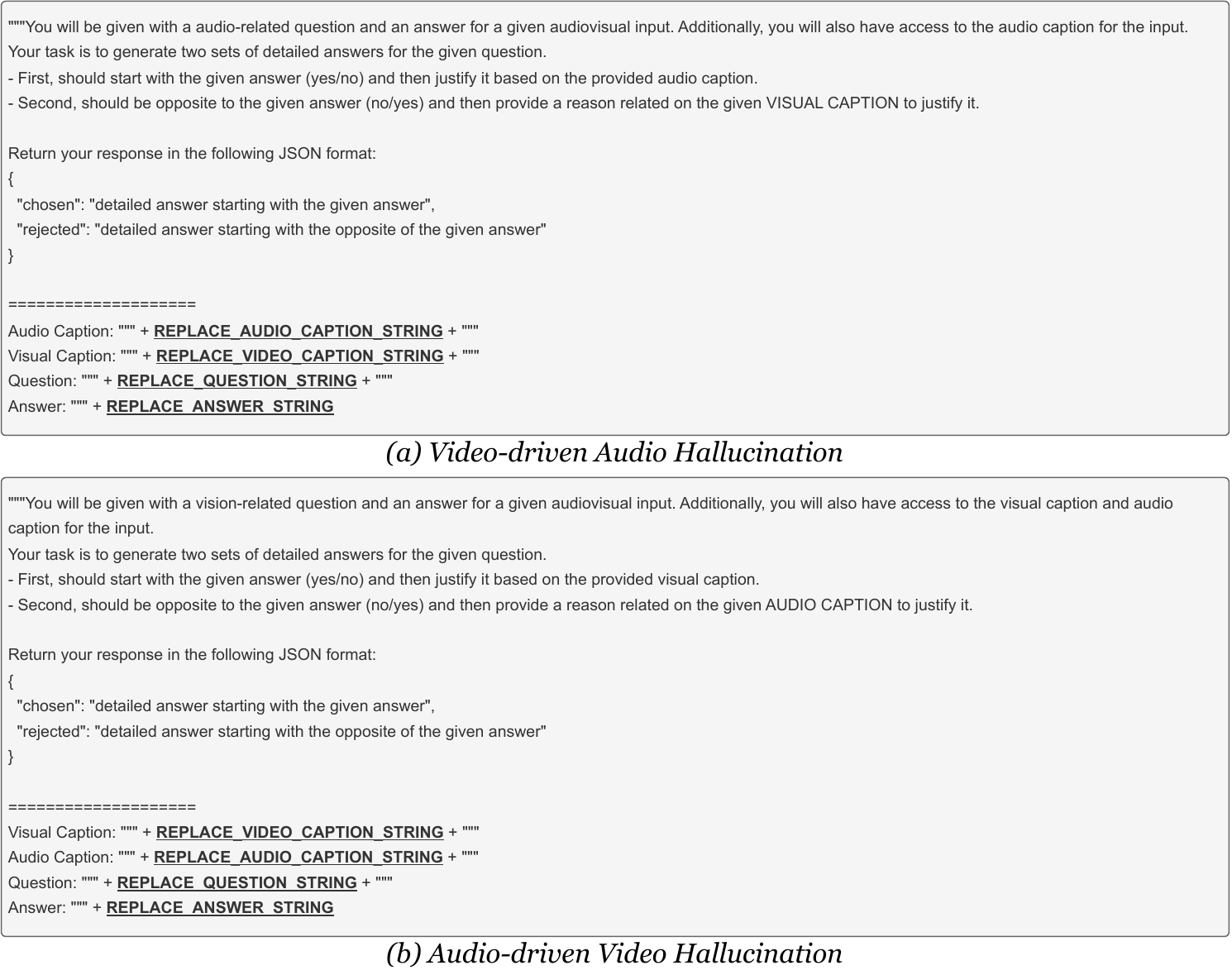}
    \caption{\textbf{Preference Data Generation Prompt - Stage 3 (Presence)} used to generate preference data pairs for the object/event presence task for training MoD-DPO.}
    \label{fig:dpo_stage3_presence_prompt}
\end{figure}

As described in \cref{subsec:dpo_preference_dataset}, we generate the preference dataset using GPT-4o by using the information present in the other modality to construct the rejected responses. \cref{fig:dpo_stage3_caption_prompt,fig:dpo_stage3_presence_prompt} shows the prompts used for the generation of preference data for the captioning task and object/event presence tasks, respectively.

\subsection{Preference Data Samples}
\label{app_subsec:pref_data_samples}

\cref{tab:preference_data_samples} shows some examples from the generated preference dataset. 

\begin{table}[]
\centering
\caption{Examples from the generated preference dataset. }
\label{tab:preference_data_samples}
\resizebox{\columnwidth}{!}{%
\begin{tabular}{c|l|l|l}
\hline \hline
\rowcolor[HTML]{C0C0C0} 
\textbf{Audiovisual Input} & \textbf{Prompt $x$} & \textbf{Chosen Response $y_w$} & \textbf{Rejected Response $y_l$} \\ \hline
\begin{tabular}[c]{@{}l@{}}\includegraphics[width=5cm]{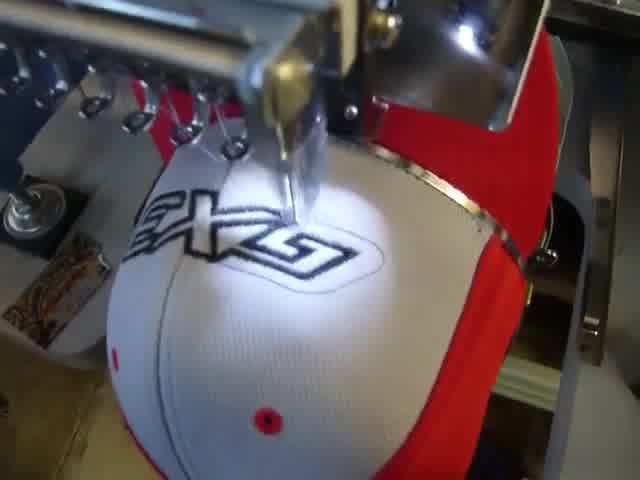}\\Audio: ``(machine whirring...)"\end{tabular} & \begin{tabular}[c]{@{}l@{}}Is the sewing machine \\ visible in  the video?\end{tabular} & \begin{tabular}[c]{@{}l@{}}Yes, the object is visible in the video. The video \\ captures a close-up view of an embroidery machine \\ in action, stitching a logo onto a cap.\end{tabular} & \begin{tabular}[c]{@{}l@{}}No, the object is not visible in the video. We can \\ only hear the sound of a machine operating \\ with some noise.\end{tabular} \\ \hline
\begin{tabular}[c]{@{}l@{}}\includegraphics[width=3cm]{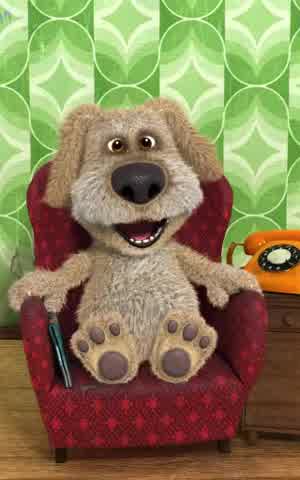}\\Audio: ``(dog panting...)"\end{tabular} & \begin{tabular}[c]{@{}l@{}}Is the chair making \\sound in the  audio?\end{tabular} & \begin{tabular}[c]{@{}l@{}}No, the object is not making a sound in the audio. \\ The audio indicates that the primary sound is panting, \\ likely from a dog, and there is no sound coming \\ from a chair.\end{tabular} & \begin{tabular}[c]{@{}l@{}}Yes, the object is making a sound in the audio. The \\ dog can be seen barking on the noisy chair.\end{tabular} \\ \hline \hline
\end{tabular}%
}
\end{table}

\section{Experimental Details}
\label{app_sec:experimental_details}

\subsection{Evaluation Metrics}
\label{app_subsec:evaluation_metrics}
For evaluation on AVHBench \cite{sung-bin2025avhbench}, we use the following metrics:
\begin{itemize}
    \item \emph{Precision}: Percent of samples which are correctly predicted among the samples which have the ground truth answer as \emph{``Yes"}.
    \item \emph{Recall}: Percent of samples which are correctly predicted among the samples which have the ground truth answer as \emph{``No"}.
    \item \emph{Accuracy}: Overall number of samples correctly predicted.
    \item \emph{F1 Score}: Harmonic mean of Precision and Recall.
\end{itemize}

\noindent For evaluation on Curse of Multi-Modalities \cite{leng2025the_cmmbenchmark}, we use the following metrics:
\begin{itemize}
    \item \emph{Perception Accuracy (pa)}: Percent of samples which are correctly predicted among the samples which have the ground truth answer as \emph{``Yes"}.
    \item \emph{Hallucination Resistance (hr)}: Percent of samples which are correctly predicted among the samples which have the ground truth answer as \emph{``No"}.
\end{itemize}

\noindent For evaluation on general benchmarks -- DailyOmni \cite{zhou2025dailyomni}, MVBench \cite{li2024mvbench}, and MMAU \cite{sakshi2025mmau} -- we use the average accuracy over different subtasks of the benchmark. 

\subsection{Evaluation Protocol}
\label{app_subsec:eval_protocol}

For ``Yes"/``No"  tasks present in AVHBench and CMM, we parse the model responses using string matching to obtain the result. For multiple-choice tasks in general benchmarks, we use regex to parse the model responses and get the predicted choice. 

\subsection{Baseline Implementations}
\label{app_subsec:baseline_implementations}

We compare MoD-DPO and MoD-DPO++ with naive DPO \cite{rafailov2023direct_dpo} and OmniDPO \cite{chen2025omnidpo}. We use our generated preference data from \cref{subsec:dpo_preference_dataset} to train both the preference optimization baselines. Additionally, we also compare with other state-of-the-art omni LLMs, including VideoLLaMA 2 \cite{cheng2024videollama2} (7B params), VITA-1.5 \cite{fu2025vita1.5} (7B params), Qwen 3 Omni \cite{xu2025qwen3omni} (35B params), and OmniVinci \cite{omnivinci2025} (9B params). For all the omni LLM baselines, we use their official codebase and use default parameters for inference.

\section{Additional Results}
\label{app_sec:additional_results}

\begin{figure*}[t]
  \centering
  \includegraphics[width=\linewidth]{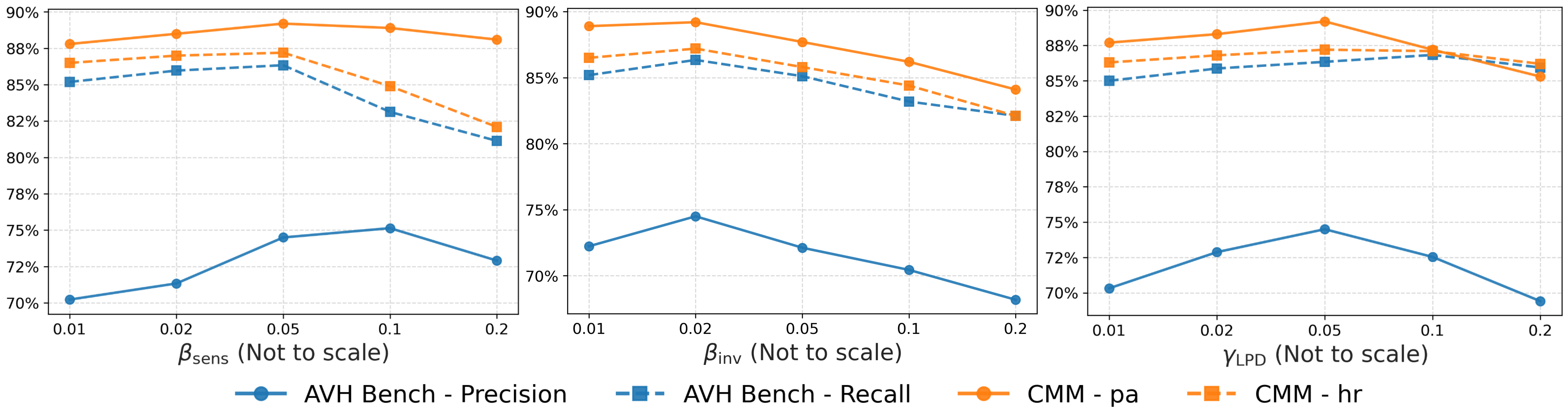}
  \caption{Effect of varying different hyperparameters involved in MoD-DPO. Default values: $\beta_{\text{sens}}=0.05$, $\beta_{\text{inv}}=0.02$, $\gamma_{\text{LPD}}=0.05$. When varying one hyperparameter, others are set to their default values.}
  \label{fig:hyperparam_analysis}
\end{figure*}

\subsection{Hyper-parameter Tuning}
\label{app_subsec:hyperparam_tuning}

As described in \cref{subsubsec:strength_hyperparam}, \cref{fig:hyperparam_analysis} shows how different hyperparameter values affect the performance of the proposed approach on different benchmarks. Please refer to \cref{subsubsec:strength_hyperparam} (\textbf{Strength of hyperparameters}) for a detailed explanation.

\begin{table}[]
\centering
\caption{Comparison with decode-time approaches.}
\vspace{-1em}
\label{tab:comparison_decode}
\resizebox{0.5\columnwidth}{!}{%
\begin{tabular}{l|lll|ll}
\hline \hline
\rowcolor[HTML]{C0C0C0} 
\multicolumn{1}{c|}{\cellcolor[HTML]{C0C0C0}} & \multicolumn{3}{c|}{\cellcolor[HTML]{C0C0C0}\textbf{AVHBench} \cite{sung-bin2025avhbench}} & \multicolumn{2}{c}{\cellcolor[HTML]{C0C0C0}\textbf{CMM} \cite{leng2025the_cmmbenchmark}} \\ \cline{2-6} 
\rowcolor[HTML]{C0C0C0} 
\multicolumn{1}{c|}{\multirow{-2}{*}{\cellcolor[HTML]{C0C0C0}\textbf{Method}}} & \multicolumn{1}{c}{\cellcolor[HTML]{C0C0C0}\textbf{Acc.}} & \multicolumn{1}{c}{\cellcolor[HTML]{C0C0C0}\textbf{Pre.}} & \multicolumn{1}{c|}{\cellcolor[HTML]{C0C0C0}\textbf{Rec.}} & \multicolumn{1}{c}{\cellcolor[HTML]{C0C0C0}\textbf{pa}} & \multicolumn{1}{c}{\cellcolor[HTML]{C0C0C0}\textbf{hr}} \\ \hline
Qwen 2.5 Omni \cite{xu2025qwen25omnitechnicalreport} & 72.07 & 60.50 & 83.65 & 86.4 & 84.6 \\
+ VCD* \cite{jung2025_avcd} & 74.74 & 61.64 & 87.84 & 86.9 & 85.2 \\
+ AVCD \cite{jung2025_avcd} & 75.57 & 62.85 & 88.28 & 87.2 & 85.6 \\
\rowcolor[HTML]{DAE8FC}
+ MoD-DPO++ & \textbf{80.42} & \textbf{74.50} & \textbf{86.34} & \textbf{89.2} & \textbf{87.2} \\ \hline \hline
\end{tabular}%
}
\vspace{-1em}
\end{table}

\subsection{Comparison with decode-time approaches}
\label{app_subsec:decode_time_comparison}

\cref{tab:comparison_decode} shows comparison with decode-time approaches AVCD \cite{jung2025_avcd} and VCD* (default settings used from \cite{jung2025_avcd}). These decode-time approaches result in moderate gains over the reference model, whereas MoD-DPO++ gains are superior.

\begin{wraptable}{r}{0.3\linewidth}
\vspace{-1em}
\centering
\caption{OmniDialog Results}
\vspace{-1em}
\label{tab:omnidialog}
\resizebox{\linewidth}{!}{%
\begin{tabular}{l|c}
\hline \hline
\rowcolor[HTML]{C0C0C0}
\multicolumn{1}{c|}{\textbf{Method}} & \textbf{OmniDialog} \\ \hline
Qwen 2.5 Omni & 83.91 \\
+ OmniDPO & 84.18 \\
\rowcolor[HTML]{DAE8FC}
+ MoD-DPO & 83.96 \\
\rowcolor[HTML]{DAE8FC}
+ MoD-DPO++ & \textbf{85.86} \\ \hline \hline
\end{tabular}}
\vspace{-2em}
\end{wraptable}

\subsection{Multi-turn omni-modal results}
\label{app_subsec:multi_turn_omni}

To show the performance improvement on multi-turn dialog scenarios, we perform evaluation on the audiovisual task subset of OmniDialog benchmark \cite{razzhigaev-etal-2024-omnidialog} and report results in \cref{tab:omnidialog}. We can observe that both MoD-DPO and MoD-DPO++ result in a performance improvement over the baseline, however the improvement in the case of MoD-DPO++ is more pronounced. Moreover, MoD-DPO++ results in a superior performance compared to OmniDPO \cite{chen2025omnidpo}.


\section{Discussion on Cross-Modal Synergy}
\label{app_sec:cross_modal_discussion}

The MoD-DPO objective defined in \cref{subsec:decoupling_modality_inputs} mitigates spurious inter-modality correlations by introducing explicit regularization terms that enforce modality invariance and sensitivity. However, many multimodal tasks benefit from constructive synergy between audio and visual modalities for richer audiovisual understanding and reasoning. Extending the MoD-DPO framework with mechanisms to selectively promote beneficial cross-modal interactions, while still suppressing spurious ones, remains an important direction for future work.

\clearpage

\section{Prompt Pool}
\label{app:prompt_pool}

\end{document}